\documentclass{article}

\usepackage{PRIMEarxiv}

\usepackage[utf8]{inputenc} 
\usepackage[T1]{fontenc}    
\usepackage{hyperref}       
\usepackage{url}            
\usepackage{booktabs}       
\usepackage{amsfonts}       
\usepackage{nicefrac}       
\usepackage{microtype}      
\usepackage{lipsum}
\usepackage{fancyhdr}       
\usepackage{graphicx}       
\graphicspath{{media/}}     

\usepackage{amsmath}    
\usepackage{multirow}
\usepackage{array}
\usepackage{makecell}   
\usepackage{bm}

\title{Robust Salient Object Detection on Compressed Images \\ Using Convolutional Neural Networks
}

\author{
  Guibiao Liao \\
  Peking University \\
  \texttt{gbliao@stu.pku.edu.cn} \
   \And
  Wei Gao \thanks{\textit{Corresponding author}}\\ 
  Peking University \\
  \texttt{gaowei262@pku.edu.cn} \\
}

\begin{document}
\maketitle

\begin{abstract}
Salient object detection (SOD) has achieved substantial progress in recent years. In practical scenarios, compressed images (CI) serve as the primary medium for data transmission and storage. However, scant attention has been directed towards SOD for compressed images using convolutional neural networks (CNNs). 
In this paper, we are dedicated to strictly benchmarking and analyzing CNN-based salient object detection on compressed images.
To comprehensively study this issue, we meticulously establish various CI SOD datasets from existing public SOD datasets. Subsequently, we investigate representative CNN-based SOD methods, assessing their robustness on compressed images (approximately 2.64 million images).
Importantly, our evaluation results reveal two key findings: 1) current state-of-the-art CNN-based SOD models, while excelling on clean images, exhibit significant performance bottlenecks when applied to compressed images. 2) The principal factors influencing the robustness of CI SOD are rooted in the characteristics of compressed images and the limitations in saliency feature learning. 
Based on these observations, we propose a simple yet promising baseline framework that focuses on robust feature representation learning to achieve robust CNN-based CI SOD.
Extensive experiments demonstrate the effectiveness of our approach, showcasing markedly improved robustness across various levels of image degradation, while maintaining competitive accuracy on clean data. 
We hope that our benchmarking efforts, analytical insights, and proposed techniques will contribute to a more comprehensive understanding of the robustness of CNN-based SOD algorithms, inspiring future research in the community. 
\end{abstract}


\section{Introduction}\label{Introduction}
Salient object detection (SOD) aims at capturing the visually interesting object regions within images \cite{gao2023pp8k, gao2023thorough, liao2024rethinking}, serving as a crucial pre-processing step for various computer vision tasks, such as compression \cite{guo2009novel, yuan2023openfastvc, gao2023opendmc, gao2020low, wu2021deep, gao2022openhardwarevc, zhang2022tdrnet, tao2023adanic}, image captioning \cite{fang2015captions} and segmentation \cite{han2005unsupervised, liao2024vlm2scene, liao2024ov}. 
In the era of explosive multimedia data growth, image compression is an essential technology for conserving transmission bandwidth and hardware storage. 
However, in the real world, the acquisition of compressed images is confronted by the constraints of limited and fluctuating bandwidth, resulting in the acquisition of images with varying degrees of degradation. 
Consequently, the development of a robust model for compressed image salient object detection (CI SOD) emerges as an imperative concern.

\begin{figure*}
\centering
\includegraphics[width=\linewidth]{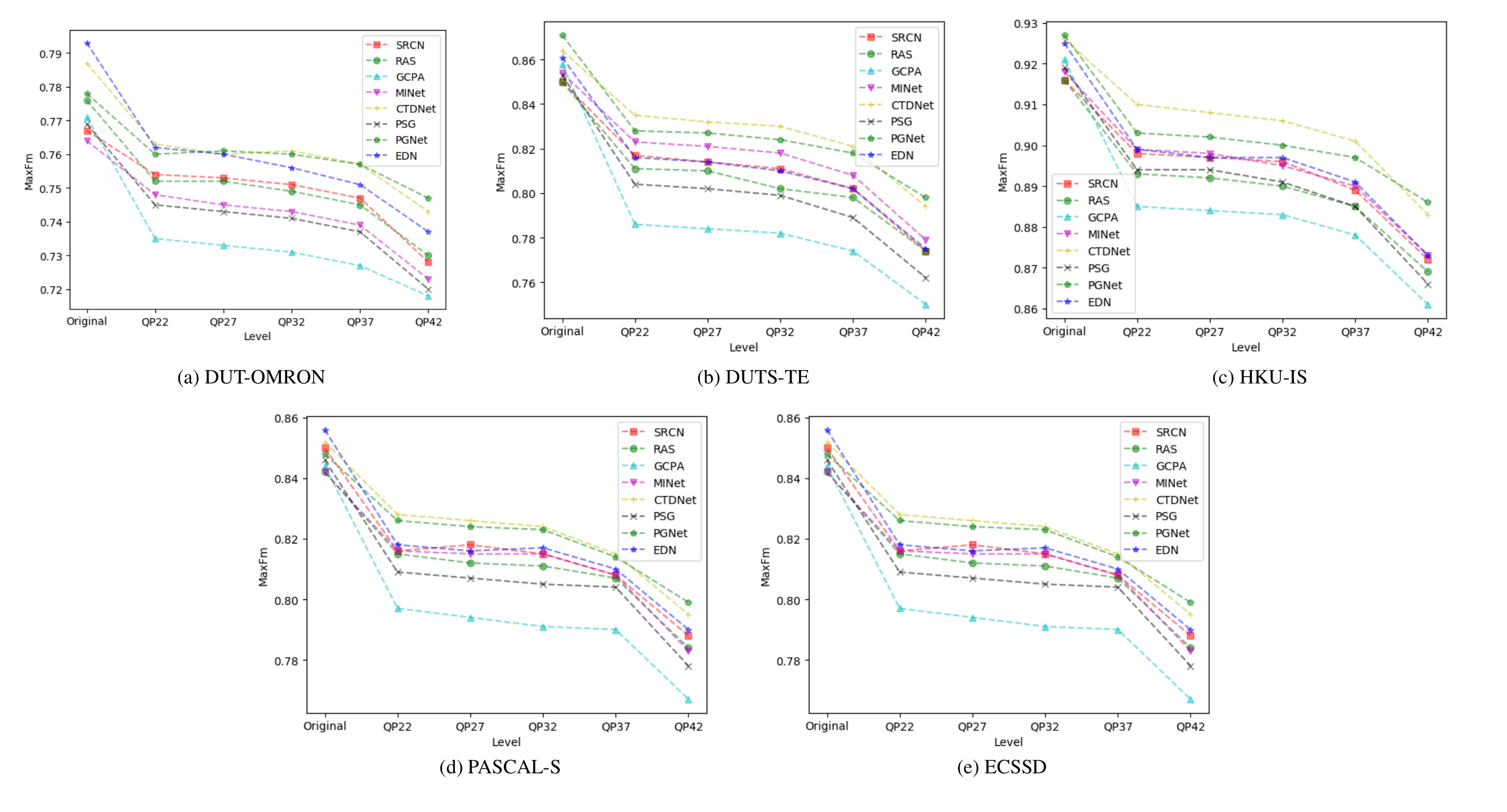}
\caption{
Performances of representative CNN-based SOD methods on different datasets, including the original test results from models trained on the original clean image dataset (i.e., Original), and the compressed test results from models retrained on corresponding compressed image datasets (i.e., QP22, QP27, QP32, QP37, QP42). 
Here, QP means the quantization parameter and higher QP represents severer degradation in images. 
From the above benchmark results, it can be seen that 
1) current SOTA models suffer from large performance bottlenecks on compressing images, although they have previously achieved great performance on clean images. 
2) As the compression distortion increases (i.e., from QP22 to QP42), the performance of the model gradually decreases. 
}
\label{fig:benchmark_results}
\end{figure*}

\begin{figure*}
\centering
\includegraphics[width=0.85\linewidth]{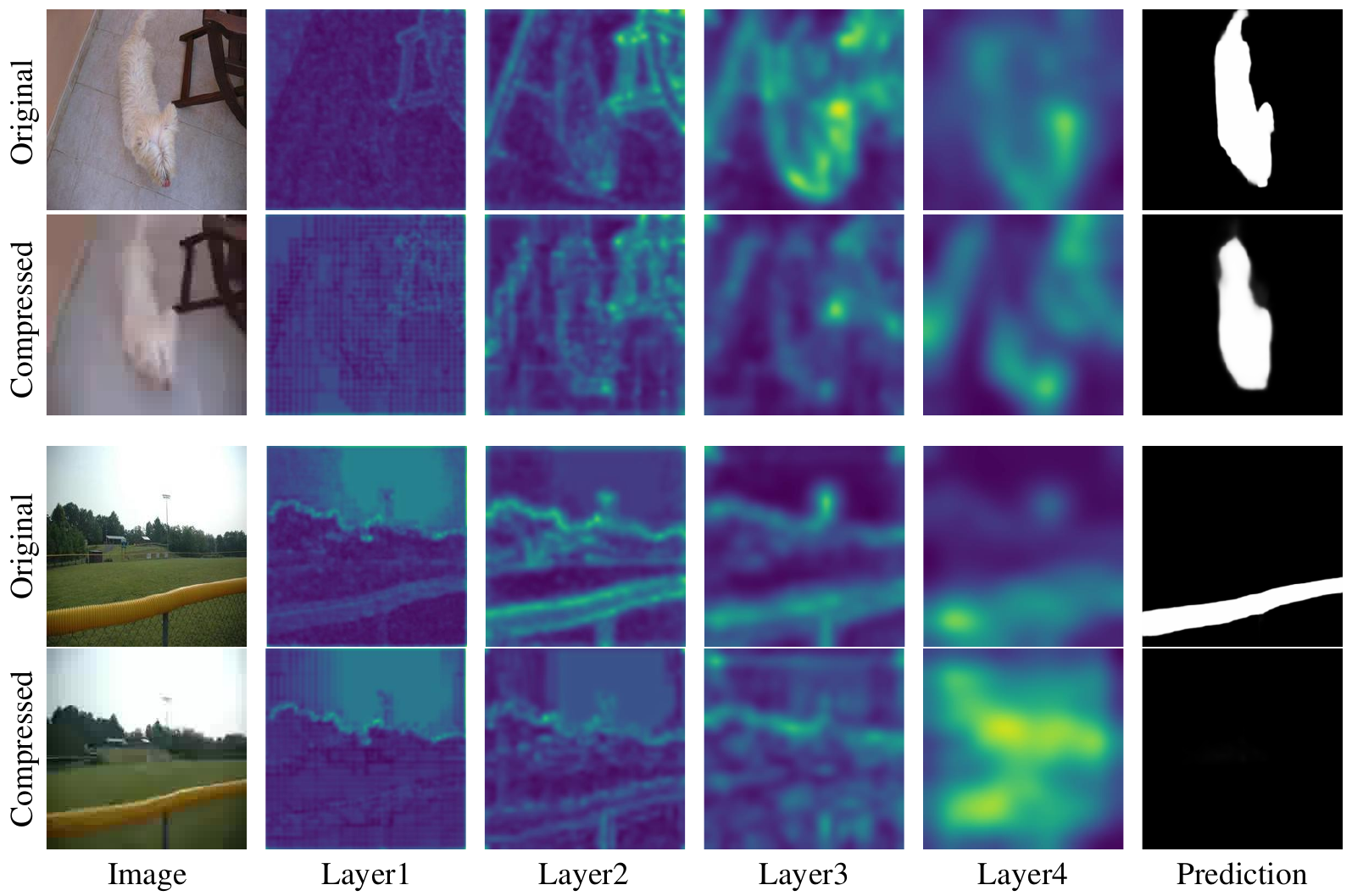}
\caption{Visualization for the hierarchical feature maps of the feature extractor and predicted saliency results for representative EDN \cite{wu2022edn}. 
It can be seen that the salient regions of the original clean image are more highlighted and accurate saliency results can be achieved. 
In contrast, the broken structure and blurring characteristics of compressed images pose great difficulties for contextual understanding and salient region detection, leading to poor results.  
}
\label{fig:Visualization}
\end{figure*}

Most existing CNN-based SOD methodologies delve into attention mechanism \cite{zhao2019pyramid,feng2019attentive}, multi-scale feature aggregation \cite{chen2020global,pang2020multi}, and boundary cue \cite{zhao2019egnet,zhou2020interactive}, yielding commendable results in the context of clean image SOD. However, scant attention has been devoted to learning CI SOD.
In this study, we conduct a comprehensive evaluation of current representative CNN-based SOD models to scrutinize their robustness for CI SOD.
Initially, to establish various CI SOD datasets, we perform image compression with five different degrees on the existing popular SOD datasets, generating C-DUT-OMRON, C-DUTS, C-HKU-IS, C-PASCAL-S, and C-ECSSD.
Subsequently, the SOD models are retrained using the compressed training benchmark, and their performances are rigorously assessed on the aforementioned CI SOD test datasets (for additional details, refer to Section \ref{Experimental Setup}).
Illustrated by the benchmark results presented in Fig. \ref{fig:benchmark_results}, our findings reveal two crucial observations.
1) First, although previous progress of SOD has led to improvements in clean images over time, their performances may not be guaranteed for compressed images, indicating a lack of inherent robustness.
2) Second, as compression distortion intensifies from left to right along the horizontal axis, the performance of SOD models progressively deteriorates.
These results underscore the imperative need to explore robust algorithmics in the realm of CI SOD.

Based on the obtained observations, we draw insights into the distinct challenges encountered in CI SOD when compared to the conventional CNN-based SOD task on clear images. Notably, two critical characteristics of compressed images significantly impact the algorithm's robustness. 
1) \textbf{Structural Discontinuity (Pixel Shifting) and Quality Unsmoothness.}
This phenomenon primarily stems from blocking artifacts introduced during the encoding process \cite{sullivan2012overview}. The lossy compression can make different blocks have different level distortions, and then the prediction and reconstruction for block-level residuals cannot guarantee the consistency of junctions, as well as the quality smoothness among blocks. Thus, the discontinuity and heterogeneity may pose a challenge to accurate CI SOD.
2) \textbf{Missing Information or Blurring of Saliency-Related Regions.}
The perceptual redundancy removal via quantization causes the loss of high-frequency information, i.e., the blurring effect, and meanwhile the limited bit rate constraint results in the inevitable pixel or block copying from the distorted background.
Hence, existing models may exhibit sensitivity to these missing or blurring effects in compressed images, potentially leading to the failure to generate complete and accurate saliency results.

To substantiate our insights, we perform a visual analysis of hierarchical feature maps from the feature extractor and predicted saliency results generated by a representative SOD model.
As illustrated in Fig. \ref{fig:Visualization}, our observations reveal two critical aspects:
1) The presence of broken structures in compressed images, disrupts spatial relationships and introduces challenges to accurate scene understanding.
2) The blurring characteristic, wherein salient regions receive diminished attention, and the intrusion of harmful noise adversely affects detection results.
These findings demonstrate the noteworthy impairments introduced by the compression process, 
emphasizing the imperative of robust feature learning for CI SOD.
Therefore, our work is dedicated to addressing this challenging but less explored CI SOD issue.
Motivated by the analysis of compressed image characteristics and the insights gained from feature visualization, we propose a simple yet potential baseline framework, which mainly concentrates on robust feature representation learning to achieve robust CI SOD.

Compared to feature learning of clear images, the structural discontinuity and blurring regions of the compressed image affect robust feature learning. Mitigating the adverse effects of these characteristics is pivotal for achieving robust CI SOD. 
Inspired by the features from clean images show more informative and accurate representation power, as illustrated in Fig. \ref{fig:Visualization}, we propose to make feature learning of compressed images imitate the feature learning of clear images, including custom \textit{relation} attribute and \textit{location} attribute. 
In essence, if we can properly leverage this feature-based mimicking to learn valuable relation and location information from clean images, the feature learning defects of compressed images can be remedied and the robustness can be improved. 
From this point of view, we present Hybrid Prior Learning (HPL), which treats features from clean images as prior knowledge and performs a hybrid feature-based distillation approach.
As depicted in Fig. \ref{fig:overall_network}, our HPL mainly comprises a hybrid prior generator and a target network. Diverse and valuable knowledge from the hybrid prior generator is transferred to the target network through three distinct strategies, elucidated as follows.

{\textbf{First}}, considering the inherent discontinuity of compressed images, we propose a relation prior learning strategy. The hybrid prior generator captures relation features from clean images that inherit both short-range and long-range spatial relation information as the feature guidance. This relation knowledge is then seamlessly transferred to the target network, facilitating the extraction of internal correlations specific to compressed images.
{\textbf{Second}}, to handle missing information and blurring regions, we introduce a location prior learning strategy. Leveraging salient foreground-guided region responses from the hybrid prior generator, this strategy enhances the localization of salient objects, providing valuable feature guidance for the target network.
{\textbf{Third}}, to synergistically exploit the learning potential of the previously two outlined strategies, we propose a self-masked learning strategy. This strategy masks the pixels of the salient region on compressed images, compelling the target network to better learn relation and location priors to generate robust features throughout the above two strategies. The generated features exhibit enhanced robustness, thereby improving the overall robustness of feature learning in our model.
Particularly, the proposed hybrid prior generator is exclusively adopted to train the target network during training, rendering it \textit{computationally free} during inference.

{Furthermore}, recognizing the significance of modeling correlations between the localization and object parts of salient objects for comprehensive CI SOD, we propose a location-aware graph method. This method explicitly models and reasons about the relationships between localization and object parts, enhancing the understanding of their inherent connections.
Overall, the main contributions of this work are summarized as follows:

\begin{itemize}

\item We present comprehensive benchmarks ($\sim$2.64 million images) for studying the challenging yet underexplored compressed image salient object detection (CI SOD) issue. Our benchmarking efforts reveal that existing CNN-based SOD models suffer from large performance bottlenecks, primarily attributed to the distinctive characteristics and deficient feature learning of compressed images. 

\item Motivated by our observations, we introduce a hybrid prior learning strategy (HPL) tailored to place emphasis on robust feature learning to improve the CI SOD robustness. HPL encourages the target network to delve into internal correlations and essential saliency-related information on compressed images, particularly addressing the challenges posed by the broken structure and missing information inherent in compressed images. 

\item Extensive experiments on various benchmarks demonstrate the superiority of our approach over state-of-the-art methods. Moreover, our approach exhibits enhanced robustness under varying degradation levels, while maintaining competitive accuracy on clean data.

\end{itemize}

The remainder of this paper is structured as follows. In Section \ref{RelatedWork}, we review and discuss the related works. Section \ref{Methodology} introduces our proposed approach in detail. Section \ref{Experiments} includes evaluation benchmarks, experimental setup, performance comparison, and ablation discussion. Finally, we conclude this work in Section \ref{Conclusion}.

\section{Related Work}\label{RelatedWork}
\subsection{Salient Object Detection}
Numerous SOD approaches \cite{he2015supercnn,wang2017salient,liu2019simple,wu2019cascaded,chen2021cnn,liao2020mmnet,wei2020f3net,tian2022learning,pang2020multi,zhao2020suppress,sun2022learnable,gao2021unified,li2023delving,liao2022cross} have been presented over the years and have shown encouraging saliency results on various benchmark datasets. 
These current SOD algorithms can roughly be divided into multi-level-based, boundary-based, and attention-based SOD approaches.

\textbf{Multi-level-based.} 
For multi-level or multi-scale modeling, some existing SOD models propose to aggregate the features of different layers from backbones to obtain a saliency result. It helps infer the saliency of various object sizes. 
Specifically, 
Chen et al. \cite{chen2020global} proposed a feature interweaved aggregation module to take advantage of high-level, low-level, and global features for more complete saliency results. 
Li et al. \cite{li2020cross} first utilized a cross-layer feature aggregation module to integrate multi-level information, and then designed the cross-layer feature distribution module to allocate previous integrated features to corresponding layers for better fusion.
Pang et al. \cite{pang2020multi} designed aggregate interaction modules and self-interaction modules to integrate adjacent layers features and adaptively deal with different scale information. 
Wu et al. \cite{wu2022edn} built a scale-correlated pyramid convolution in the decoder for better multi-level features fusion. 
In ICON \cite{zhuge2022salient}, Zhuge et al. introduced a diverse feature aggregation module to integrate features from various layers.  
Liu et al. \cite{liu2022poolnet+} first established a global guidance module and designed a feature aggregation module at different feature levels to integrate multi-level features with the guidance of the global guidance module.

\textbf{Boundary-based.}
Boundary-guided learning is another commonly used mechanism in SOD. Most current boundary-based SOD methods focus more on the object contours by adopting additional boundary learned loss functions. 
For example, Zhao et al. \cite{zhao2019egnet} treated the salient object edge information with various resolutions as complementary information to help locate the salient regions. 
In \cite{qin2019basnet}, a hybrid loss for boundary-aware saliency feature learning is used to predict the object with fine boundaries in pixel, patch, and map levels. 
Zhou et al. \cite{zhou2020interactive} regarded the pixels near the boundary as hard examples and proposed an adaptive contour loss to adaptively discriminate these hard examples during training. 
PAGE \cite{wang2019salient} proposed a salient edge detection module to highlight explicit salient edge information for better salient objects locating with sharping boundaries. 
SRCN \cite{wu2019stacked} proposed an edge-aware model to interactively pass information between the saliency map and edge map. 
In BANet \cite{su2019selectivity}, a boundary-aware network is designed to incorporate a boundary localization stream for salient boundaries with high selectivity detecting. 
Zhao et al. \cite{CTDNet21} introduced a boundary path to explicitly utilize boundary information for boundary quality improvement.

\textbf{Attention-based.}
In order to effectively capture salient objects, a group of approaches enhances feature learning by paying more attention to useful features.  
In \cite{zhao2019pyramid}, Zhao et al. proposed a pyramid feature attention approach, which is designed to enhance multi-level features by channel and spatial attention.
Feng et al. \cite{feng2019attentive} built attentive feedback modules between the encoder and decoder, which aims to capture the overall structure of objects. 
Chen et al. \cite{chen2020reverse} designed a reverse attention block to perform an attention fusion operation on different side-output layers. 
Liu et al. \cite{liu2021visual} employed the multi-head self-attention and patch-task-attention mechanism to perform saliency detection. 
\cite{lee2022tracer} employed a masked edge attention module and a union attention module in the encoder and decoder, respectively. The former was used to propagate the refined edge information and the latter was applied to aggregate complementary channel and important spatial information. 
\cite{hussain2022pyramidal} applied a residual convolutional attention decoder to conduct pyramidal attention manner for fine-grained saliency prediction generation. 
In \cite{xie2022pyramid}, an attention-based cross-model grafting module and an attention-guided Loss are proposed to promote feature learning. 
Ma et al. \cite{ma2023boosting} proposed a complementary attention module to dynamically provide spacial-wise and channel-wise attention for detail and semantic features from the encoder. 
Recently, Zhou et al. \cite{zhou2020multi} proposed an attention transfer network for degraded SOD. However, it only implemented pixel-wise selections of features, which ignored the structural discontinuity and blurring characteristics of compressed images, resulting in limitations in CI SOD.

Although the aforementioned approaches can boost the SOD performance on clean images, they ignore the challenges of compressed images, which can be vulnerable in CI SOD tasks, showing less robustness on compressed images as shown in Fig. \ref{fig:benchmark_results}. 
Differently, our work focuses on the under-explored CI SOD issue, and proposes a hybrid prior learning strategy (HPL). HPL considers the spatial correlation and crucial location information on compressed images, which can better learn robust representations to achieve better CI SOD.

\subsection{Knowledge Distillation}
Knowledge distillation (KD), also known as teacher-student learning, is a commonly used method for transferring information from a larger network to a small network, aiming to improve the efficiency of deep neural models. 
Hinton et al. \cite{hinton2015distilling} first proposed the concept of KD, in which the output logits of the teacher network are transmitted to the student network to boost the image classification performance of the student network. 
after this, KD has been widely used in many fields to boost the efficiency or performance of deep neural models. 
For instance, 
in \cite{ge2018low_face}, Ge et al. used a high-resolution and high-accuracy face network as a teacher and proposed to transfer knowledge from the teacher network to the low-resolution face network for performance improvement. 
\cite{komodakis2017paying} designed an attention distillation strategy at a pixel-wise level to improve the image classification performance of the lightweight network. 
In \cite{li2017mimicking}, a feature map mimic approach was developed, which used high-level feature activation from the large model and pixel-wise mimicking to assist a small object detection network.  
Hu et al. \cite{hu2021boosting} employed L1 loss to impose supervision on the depth map knowledge learning to improve the depth estimation performance for a lightweight network. 
Farhadi et al. \cite{bajestani2020tkd} transferred the temporal knowledge over the selected video frame from a larger teacher network to the lightweight student model, boosting the efficiency of video recognition with a temporal knowledge distillation strategy. 
Zhang et al. \cite{zhang2022kd} effectively improved the salient object detection accuracy by first using a bigger pre-trained teacher network to provide a saliency prediction that is used to offer weak label supervision for an untrained tiny model.

However, in most of these methods, the student model is usually trained by distilling the individual pixel information. Such individual distillation neglects the spatial relationships between features and the importance of different features on compressed images, but such information plays an important role in CI SOD. 
In contrast, we propose the HPL take into account spatial correlation and crucial features rather than just pixel-wise. Thus, we design tailored relation prior and location prior learning strategies, which explicitly exploit valuable relationship and salient region information, making our model suitable and more robust to broken structure and blurring areas of compressed images.

\section{Methodology}\label{Methodology}
In this section, we first introduce the overview of our proposed methodology in Sect. \ref{Methodology_Overview}. Then, we sequentially give a detailed explanation for our Hybrid Prior Learning (HPL) in Sect. \ref{Methodology_HPL}, and Location-aware Graph Reasoning (LGR) in Sect. \ref{Methodology_LGR}. Finally, we illustrate the overall training objective of our approach in Sect. \ref{Methodology_Loss}.

\begin{figure*}[!t]
\centering
\includegraphics[width=\linewidth]{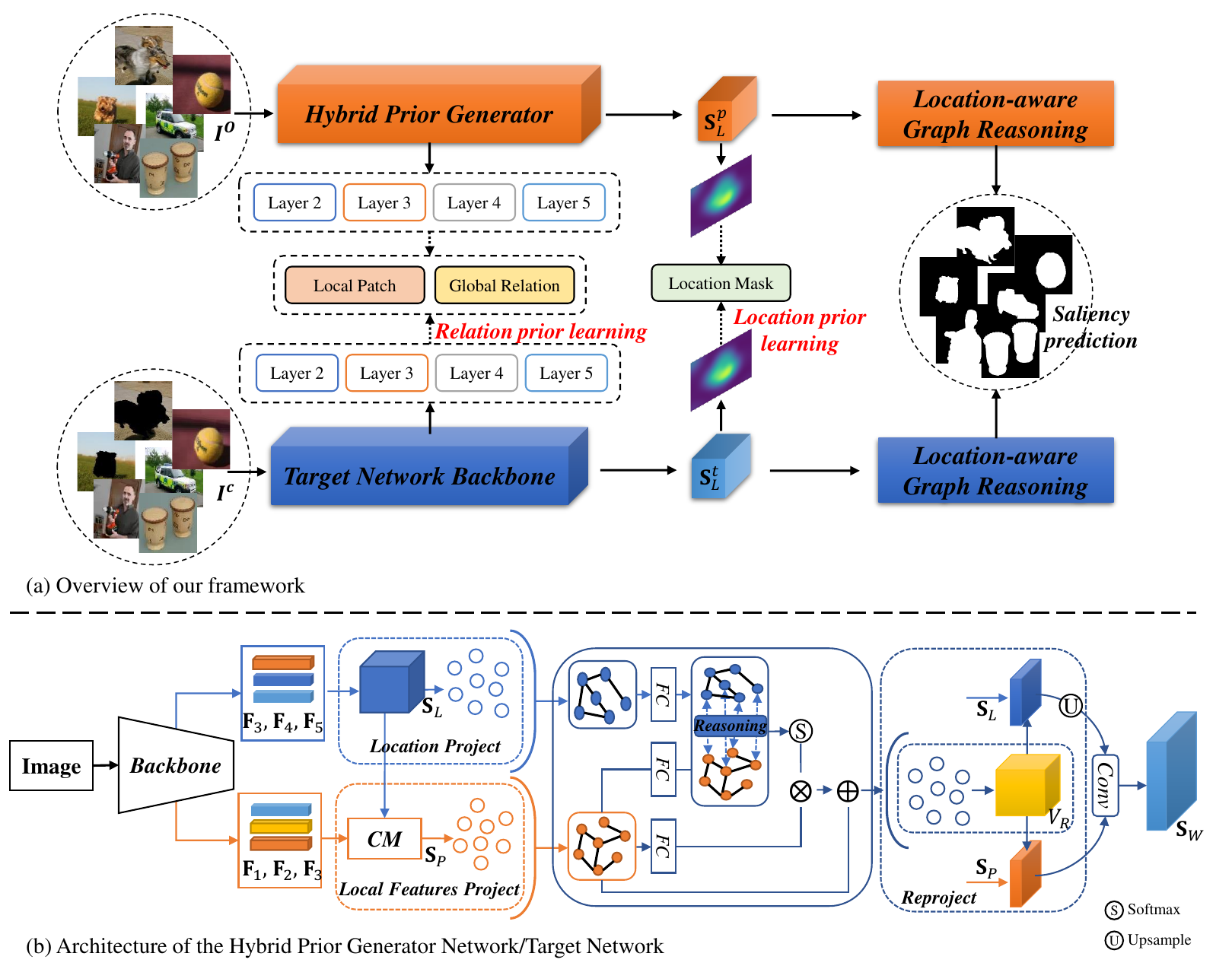}
\caption{The overall architecture of our proposed approach. 
(a) Illustration of our framework. 
Particularly, the Hybrid Prior Generator network is only adopted to train the target network during training, and it is omitted during inference. 
(b) The overall architecture of the Hybrid Prior Generator network and Target Network, both of which have the same structure. }
\label{fig:overall_network}
\end{figure*}

\subsection{Overview} \label{Methodology_Overview}

\textbf{Motivation.} 
As aforementioned, the broken structure and blurring characteristic on compressed images introduce noteworthy impairments to the robustness of algorithms and show the importance of robust feature learning for CI SOD. 
Based on the aforementioned observations, we find that 
1) as shown in Fig. 2, the features from clean images show more informative and accurate representation power.  
2) Compared to feature learning of clear images, the structural discontinuity and blurring regions of the compressed image affect robust feature learning. 
Therefore, we propose to leverage feature-based mimicking to learn valuable feature representations (e.g., \textit{relation} and \textit{location} information) from clean images, improving robust feature representation of compressed images.

\textbf{Overall Structure.} 
The architecture of our proposed framework is illustrated in Fig. \ref{fig:overall_network}. 
It mainly consists of a hybrid prior generator and a target network.
These two networks have the same network structure, but their inputs are different. 
As aforementioned, the target network is designed for detecting salient objects on compressed images. The hybrid prior generator aims to provide prior knowledge generated from clean images. 
To be specific, we denote $I^o$ as the original clean RGB image, $I^c$ as the corresponding compressed RGB image, and $y$ as the pixel-wise salient object label. 
The training dataset can be defined as $\{ I^o_i, I^c_i, y_i \}_{i=1}^N$, where $N$ is the size of the training dataset. We first use $\{I^o_i, y_i \}_{i=1}^N$ to train the hybrid prior generator for prior knowledge generation. Then, we use $\{I^c_i, y_i \}_{i=1}^N$ to train the target network affected by the priors from the hybrid prior generator. 
Through this Hybrid Prior Learning (HPL) strategy, the target network can improve robust feature representation and achieve better saliency perception on compressed images.  
Then, benefiting from the proposed HPL strategy, we introduce a Location-aware Graph Reasoning (LGR) strategy to further detect complete objects. 
Details of HPL and LGR will be presented in the following subsections.

\subsection{Proposed Hybrid Prior Learning} \label{Methodology_HPL}
Exploring the accurate feature representations is essential to detect salient regions on compressed images. 
Based on the analysis of the characteristics of compressed images and the observation of feature visualization, we propose a simple yet potential hybrid prior learning (HPL) strategy for robust CI SOD. 
As illustrated in Fig. \ref{fig:overall_network}, our HPL mainly consists of a hybrid prior generator and a target network, and diverse useful knowledge from the hybrid prior generator is transferred to the target network through \textit{relation prior learning}, \textit{location prior learning}, and \textit{self-masked learning}, respectively. 
The relation prior learning (RPL) strategy and the location prior learning (LPL) strategy are proposed to handle the discontinuity and blurring regions on compressed images, respectively. 
In addition to mimicking features of the prior generator, a self-masked learning (SML) strategy is proposed to force our target network to generate the feature of the prior generator, bringing the target network further improvements for CI SOD.

\textbf{Relation Prior Learning (RPL).} 
The vanilla feature-based KD usually adopts $l_1$-norm or $l_2$-norm to minimize the pixel-to-pixel feature error between teacher and student. 
However, due to the discontinuity and quality unsmoothness caused by blocking artifacts \cite{sullivan2012overview} on compressed images, the most widely used manner may be sub-optimal for CI SOD. 
Because it conducts individual feature learning while neglecting spatial relationships. 
To consider spatial dependencies, we therefore change the feature learning from pixel-wise to relation-wise and we propose a relation prior learning scheme. It promotes the target network to mine the short-range and long-range relationship on compressed images, with the guidance of the hybrid prior generator on clean images. 
It is further able to take into account the relationship between the target network and the prior generator, rather than just pixel-wise. 
Specifically, we use $\{\mathbf{F}_i^p\}_{i=1}^5 \in \mathbb{R}^{C\times H \times W}$ and $\{\mathbf{F}_i^t\}_{i=1}^5 \in \mathbb{R}^{C\times H \times W}$ to denote the layer-$i$ features of the backbones in the prior generator and the target network, respectively.

First, inspired by the success of SSIM \cite{wang2004image} in measuring structural similarity, we propose to leverage SSIM as a short-range relation learning operator to provide an elegant way for local patch-wise structure imitation. For each local patch, we compute the mean, variance, and cross-correlation properties to acquire the relationship between $\mathbf{F}^p$ and $\mathbf{F}^t$ by inheriting the modeling of SSIM. Thus, we force the target network to mimic the feature map from the prior generator with the local patch-wise structure guided for effective short-range relation learning, which is formulated as:  
\begin{equation}
\mathcal{L}_{SR} = \sum_{i=2}^{5} 
( \mathcal{L}_{\xi}(\phi^t_i(\mathbf{F}^t_i), \phi^p_i(\mathbf{F}^p_i))  ), 
\end{equation}  
where $\phi^t_i$ and $\phi^p_i$ are the adaptation layers, which are designed to promote the training stability by using 1 $\times$ 1 convolution. $\mathcal{L}_{\xi}$ is the SSIM measurement function and we use a Gaussian-weighted patch of size 11 $\times$ 11 to compute the local patch-wise information. 
The proposed $\mathcal{L}_{SR}$ evaluates the short-range relation difference between the target network and the prior generator feature spaces.

Second, we not only consider the short-range spatial relationship but also take into account the long-range spatial dependencies to effectively enhance the relation prior learning. 
For the layer-$i$ feature from the backbone, we could calculate its inter-pixel spatial relationship between different positions to form a relation matrix $\mathbf{R}$, which can be implemented as: 
\begin{equation}
\mathbf{R}_{i}(u, v) = \Phi( \theta_i (\mathbf{F}_i(u)), \psi_i (\mathbf{F}_i (v)) ), 
\end{equation}
where $\Phi$ denotes the scaled product operation. 
$\theta$ and $\psi$ indicate 1 $\times$ 1 convolution layers. 
$u$ and $v$ represent the spatial positions of the feature map. 
Furthermore, we assign more importance to relative features by formulating $\sigma(\mathbf{R}) \otimes \varphi(\mathbf{F}_i)$, achieving a relation enhanced feature map $\hat{\mathbf{F}}_i$. Here, $\sigma$ is the softmax function. $\varphi$ is a 1 $\times$ 1 convolution layer. $\otimes$ represents the element-wise multiplication. 
The higher intensity in feature map \textbf{$\hat{\mathbf{F}}^p_i$} from the prior generator corresponds to the valuable relation information. These discriminative features can be treated as prior knowledge and we enforce the target network to mimic this knowledge, which benefits the robust saliency perception.   
Therefore, we introduce the L2 loss function $\mathcal{L}_2$ to encourage the target network to learn the prior knowledge from the prior generator, and the long-range relation prior learning process can be calculated as: 
\begin{equation}
\mathcal{L}_{LR} = \sum_{i=2}^{5} 
( \mathcal{L}_2(\mathbf{R}^t_i, \mathbf{R}^p_i) +  \mathcal{L}_2(\hat{\mathbf{F}}^t_i, \hat{\mathbf{F}}^p_i) ), 
\end{equation}  
In summary, the relation prior learning consists of short-range relation prior and long-range relation: 
\begin{equation}
\mathcal{L}_{RPL} = \mathcal{L}_{SR} + \mathcal{L}_{LR}, 
\end{equation}  
which allows our target network to focus on exploiting the spatial relationship with the guidance of informative knowledge from the prior generator.

\textbf{Location Prior Learning (LPL).}
Due to the missing and blurring of saliency-related regions on compressed images, accurate localization of salient objects is required for robust CI SOD. 
Hence, we propose a location prior learning strategy to constrain the \textit{object-location} of the target network on compressed images to be consistent with the prior location knowledge on clean images.

As illustrated in Fig. \ref{fig:overall_network}, we first integrate high-level features by using a multi-level aggregation module, i.e., $\Psi = Cat[ Conv(U(\mathbf{F}_{5})), Conv(\mathbf{F}_{4}\otimes U(\mathbf{F}_{5})), Conv(\mathbf{F}_3 \otimes U(\mathbf{F}_{4}) \otimes U(\mathbf{F}_{5}) ) ]$. 
Thus, we aggregate $\mathbf{F}_3^t, \mathbf{F}_4^t, \mathbf{F}_5^t$, to generate the saliency location map $\mathbf{S}_L^t$ $\in \mathbb{R}^{\frac{H}{8} \times \frac{W}{8} \times 32}$, which can be formulated as: $\mathbf{S}_{L}^t = Conv(\Psi^t (\mathbf{F}_3^t, \mathbf{F}_4^t, \mathbf{F}_5^t))$. Similarly, the saliency location map $\mathbf{S}_L^p$ from the prior generator can be obtained: $\mathbf{S}_{L}^p = Conv(\Psi^p (\mathbf{F}_3^p, \mathbf{F}_4^p, \mathbf{F}_5^p))$. 
Here, $U$ denotes the upsample operation and $Cat$ denotes the channel concatenation. $Conv$ represents the convolution layer. $\otimes$ indicates the element-wise multiplication.  
Considering that the absolute value of each element in feature maps indicates its importance, we thus compute the statistical values of features across the channel dimension to obtain a regional saliency-related response for location prior learning. 
Moreover, to mitigate the effect of useless background regions, we further regularize the target network to mimic the location knowledge from the prior generator with an explicit foreground-guided mask $M$, which can be achieved from the ground truth. 
Therefore, the whole location prior learning can be implemented by:     
\begin{equation}
\mathcal{L}_{LPL} = \mathcal{L}_{2}( M (max_{k=1,C} (\mathbf{S}_{L,k}^t)), M (max_{k=1,C}({\mathbf{S}_{L, k}^p})) ) , 
\end{equation}    
where $C$ is the channel number of the feature map.  
$\mathbf{S}_{L,k}$ denotes the $k$-th slice of $\mathbf{S}_{L}$ in the channel dimension. 
Through foreground mask $M$ guiding, LPL can isolate the adverse effects of irrelevant background noises and highlight more informative saliency regional features.   

Besides, the saliency location map $\mathbf{S}_{L}^t$ of the target network is also supervised by the IoU loss and binary cross entropy loss $\mathcal{L}_{S}$ for saliency prediction, which is defined as:
\begin{equation}
\mathcal{L}_{sal1} = \mathcal{L}_{S}( U (Conv_{1 \times 1}(\mathbf{S}_{L}^t)), y ) , 
\label{Eq_location_sal} 
\end{equation}      
where $y$ is the ground truth and $Conv_{1 \times 1}$ is the $1 \times 1$ convolution. $U$ is the upsampling operation, which is used to upsample the saliency prediction to the size of $y$.

\textbf{Self-Masked Learning (SML).}
Benefiting from the relation prior learning and location prior learning strategies, adjacent and saliency-related perception of our target network can be improved to a certain extent with the proposed feature mimicking. 
Furthermore, to be more robust to the broken structure and incomplete saliency-related regions, we propose to conduct a feature generative learning strategy, namely self-masked learning (SML). 
SML works by masking the pixels of the salient region on compressed images, and then forcing the target network to use left pixels to generate the feature representation throughout the relation prior learning and location prior learning. 
During training, the probability of implementing the SML to a sample is empirically set to 10\% for stable improvement, which is discussed in the experiment. 
To sum up, such a strategy has the following benefits to CI SOD models. 
\begin{itemize}
    \item This approach can provide diverse samples due to the random implementation probability and different training iterations, which encourages our model to care more about robust representation learning. 

    \item Through such generative learning, our SML can further exploit the hybrid prior learning potential to restore robust features. As a result, the negative effect of broken and blurring issues of compressed images can be mitigated to a certain extent, further helping the target network obtain a robust representation.  

    \item SML is also only adopted during training without introducing extra computation and parameters during inference, which is computationally free to yield better performance of SOD models on compressed images with different degradation levels. 
\end{itemize}

Different from previous typical SOD models using a single backbone for feature learning on compressed images, it is worth noting that our proposed HPL can facilitate our target network to mine relation information and saliency-location features on compressed images, improving the robust feature learning for CI SOD.
More importantly, our HPL is only required during training, making it computationally free to promote the robustness of our approach for CI SOD.


\begin{figure}
\centering
\includegraphics[width=.6\linewidth]{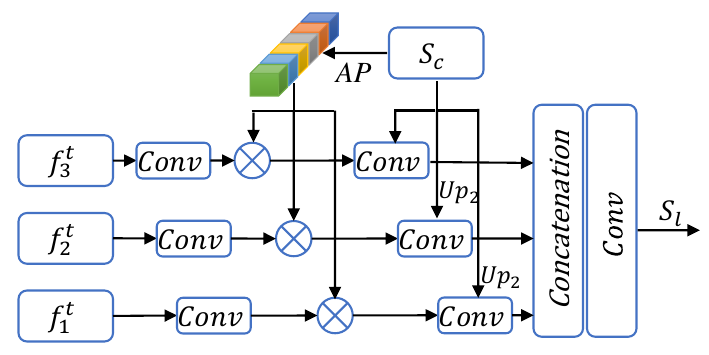}
\caption{Detailed architecture of connection module (CM).}
\label{fig:CM_module}
\end{figure}

\subsection{Proposed Location-aware Graph Reasoning} \label{Methodology_LGR}
Benefiting from the proposed HPL strategy, we can obtain more accurate object location information from the feature extractor. 
Thus, our next goal is to focus on how to capture complete and precise salient objects on compressed images. Observing the broken property of compressed images, we find that modeling the interrelation between the localization and the object parts of salient objects is critical to achieving the above goal. 
Previous works usually attempt to apply a multi-level manner \cite{pang2020multi,wu2022edn,zhuge2022salient,liu2022poolnet+} or attention mechanism \cite{liu2021visual,lee2022tracer,xie2022pyramid,ma2023boosting} for feature fusion. However, these methods do not explicitly model the relation between location and the object parts on the discontinuous and blurring compressed images, resulting in the performance bottleneck.  
To address this issue, we propose a location-aware graph reasoning (LGR) strategy, suitable for capturing the interaction of object location and local parts of the object within the challenging compressed distortion scene.
Our LGR inherently involves \textit{reasoning} from the location to its local parts, which is able to focus on the complete salient foreground for robust CI SOD.

First, we leverage the position information from the location map $\mathbf{S}_{L}^t$ to preliminarily select informative details at shallow layers as illustrated in Fig. \ref{fig:CM_module}, which can be formulated as:
\begin{equation} 
\mathbf{S}_P^t =\bm{CM}(\mathbf{S}_{L}^t, \mathbf{F}_3^t, \mathbf{F}_2^t, \mathbf{F}_1^t), 
\end{equation}       
where $\bm{CM}$ is the connection module (CM) and $\mathbf{S}_P^t$ $\in \mathbb{R}^{\frac{H}{4} \times \frac{W}{4} \times 32}$ is the object part feature expression.

Then, we conduct the graph projection operation \cite{li2018beyond} to transform $\mathbf{S}_{L}^t$ into a location graph $\mathcal{G}_L = (\mathcal{V}_L, \mathcal{E}_L)$ by using a transform matrix $\mathcal{M}_L^{\rm T}$, where the nodes $\mathcal{V}_L$ and edges $\mathcal{E}_L$ correspond to the localization and relationships respectively. 
Similarly, we build the object part feature graph $\mathcal{G}_P = (\mathcal{V}_P, \mathcal{E}_P)$ by transforming $\mathbf{S}_P^t$ with $\mathcal{M}_P^{\rm T}$.
The nodes $\mathcal{V}_P$ operate over features from local parts of the object in space. 
The edges $\mathcal{E}_P$ process messages between different nodes, i.e., different parts of the object. 
To explicitly reason the relationship between location graph representation and correlative object parts, we capture the dependencies with similarity learning. To this end, as illustrated in Fig. \ref{fig:overall_network}, we first use weight matrixes to transform $\mathcal{V}_L$ and $\mathcal{V}_P$, and then measure the correlation by using matrix multiplication. Next, we propagate the information and highlight the interrelated nodes with the correlation matrix. This process can be formulated as:  
\begin{equation}
\mathcal{\hat{V}}_P = {\rm softmax}(\rho(\mathcal{V}_P)^{\rm T} \omega(\mathcal{V}_L)) \eta(\mathcal{V}_L) + \mathcal{V}_P, 
\end{equation}    
where weight matrixes $\rho, \omega$, and $\eta$ are implemented with FCs, and we use a residual connection for better optimization. Afterward, we implement two graph convolution layers \cite{kipf2016semi} to encourage the graph connection for sufficient expression and achieve $\mathcal{{V}}_R$. 
Such a reasoning process explores correlative body parts by explicitly modeling the relationship structure and highlighting interrelated nodes and edges affected by the location graph representation. 
After the reasoning, we reprojected the enhanced graph representation $\mathcal{V}_R$ back into the original coordinate space: $\mathcal{F}_P = \mathcal{M}_P \mathcal{V}_R + \mathbf{S}_P^t$, $\mathcal{F}_L = \mathcal{M}_L \mathcal{V}_R + \mathbf{S}_L^t$. 
Here, $\mathcal{M}_P$ and $\mathcal{M}_L$ are the reprojected transform matrixes.
Then, we conduct a fusion to generate the location-aware graph reasoning output $\mathbf{S}_R^t$: 
\begin{equation}
\mathbf{S}_R^t = Conv(\mathcal{F}_P + U(\mathcal{F}_L)), 
\end{equation} 
Finally, we upsample the reasoning output $\mathbf{S}_R^t$ $\in \mathbb{R}^{\frac{H}{4} \times \frac{W}{4} \times 32}$ to the size of the ground truth, and obtain $\mathbf{S}_W^t$ $\in \mathbb{R}^{H \times W \times 1}$ with the $1 \times 1$ convolution. As a result, we can obtain $\mathcal{L}_{sal2}:$ $\mathcal{L}_{sal2} = \mathcal{L}_{S}(\mathbf{S}_W^t, y)$. 
To sum up, the cooperation of HPL and LGR can enhance the robust feature learning of our target network, improving the robustness of CI SOD.

\subsection{Overall Objective Function} \label{Methodology_Loss}
The overall objective function $\mathcal{L}_{total}^T$ for training our target network is composed of the relation prior loss $\mathcal{L}_{RPL}$, location prior loss $\mathcal{L}_{LPL}$, and saliency losses $\mathcal{L}_{sal1}$ and $\mathcal{L}_{sal2}$:
\begin{equation}
\mathcal{L}_{total}^T = \mathcal{L}_{sal1} + \mathcal{L}_{sal2} + \alpha \cdot \mathcal{L}_{RPL} + \beta \cdot \mathcal{L}_{LPL}, 
\label{eq_weight_HPL}
\end{equation}      
where the hyper-parameters $\alpha$ and $\beta$ balance the influences of different losses, and $\alpha$ and $\beta$ are empirically set to 0.5 and 0.5 in our experiments (refer to Table \ref{table_ablation_hyper_parameter}). Particularly, the objective function $\mathcal{L}_{total}^P$ for training the prior network only contains saliency losses $\mathcal{L}_{sal1}$ and $\mathcal{L}_{sal2}$.

\begin{table*}[!t]
\renewcommand{\arraystretch}{1.}
\centering
\normalsize
\caption{Performances of our approach and state-of-the-art methods (ResNet-based) using S-measure $ (S_m)$ \cite{fan2017structure}, mean absolute error $(M)$, maximum F-measure $ (F_\beta) $ \cite{achanta2009frequency}. We report the \textbf{average results} of five degradation levels. The best score of each comparison is shown in \textbf{bold}. $\uparrow$ ($\downarrow$) indicates larger (smaller) is better. Our approach can achieve very competitive performances in different settings. Particularly, the superior results under setting 1 indicate that our model shows stronger robustness to \textbf{unforeseen} compressed data when only training on the clean data. }
\scalebox{0.75}{
\begin{tabular}{c|p{0.9cm}<{\centering}|p{0.9cm}<{\centering}|p{0.9cm}<{\centering}|p{0.9cm}<{\centering}|p{0.9cm}<{\centering}|p{0.9cm}<{\centering}|p{0.9cm}<{\centering}|p{0.9cm}<{\centering}|p{0.9cm}<{\centering}|p{0.9cm}<{\centering}|p{0.9cm}<{\centering}|p{0.9cm}<{\centering}|p{0.9cm}<{\centering}|p{0.9cm}<{\centering}|p{0.9cm}<{\centering}}
\toprule[1pt]
\multirow{2}{*}{Method} 
& \multicolumn{3}{c|}{C-DUT-OMRON}  
& \multicolumn{3}{c|}{C-DUTS-TE} 
& \multicolumn{3}{c|}{C-HKU-IS}  
& \multicolumn{3}{c|}{C-PASCAL-S} 
& \multicolumn{3}{c }{C-ECSSD} \\
& $S_m \uparrow$    & $F_\beta \uparrow$   & $M \downarrow$    
& $S_m \uparrow$    & $F_\beta \uparrow$   & $M \downarrow$ 
& $S_m \uparrow$    & $F_\beta \uparrow$   & $M \downarrow$ 
& $S_m \uparrow$    & $F_\beta \uparrow$   & $M \downarrow$ 
& $S_m \uparrow$    & $F_\beta \uparrow$   & $M \downarrow$ \\
\midrule[0.8pt] %
\multicolumn{16}{c}{\normalsize Results of setting 1 (The model is trained on clean data (i.e., DUTS-TR))} \\
\midrule[0.8pt]
PoolNet \cite{liu2019simple}        & 0.772   & 0.675  & 0.068   & 0.776   & 0.709   & 0.070   & 0.822   & 0.809  & 0.065   & 0.763   & 0.727  & 0.110   & 0.838   & 0.836  & 0.075  \\
CPD \cite{wu2019cascaded}           & 0.730   & 0.642  & 0.077   & 0.704   & 0.637   & 0.091   & 0.754   & 0.750  & 0.086   & 0.688   & 0.667  & 0.142   & 0.740   & 0.761  & 0.113  \\
BASNet\cite{qin2019basnet}          & 0.758   & 0.674  & 0.073   & 0.723   & 0.653   & 0.088   & 0.774   & 0.771  & 0.079   & 0.700   & 0.667  & 0.136   & 0.770   & 0.776  & 0.099  \\
EGNet\cite{zhao2019egnet}           & 0.768   & 0.673  & 0.067   & 0.775   & 0.710   & 0.071   & 0.819   & 0.806  & 0.067   & 0.751   & 0.715  & 0.116   & 0.832   & 0.829  & 0.079  \\
SRCN \cite{wu2019stacked}           & 0.747   & 0.662  & 0.079   & 0.728   & 0.668   & 0.089   & 0.784   & 0.782  & 0.080   & 0.722   & 0.700  & 0.131   & 0.779   & 0.799  & 0.099  \\
RAS  \cite{chen2020reverse}         & 0.734   & 0.644  & 0.077   & 0.705   & 0.631   & 0.091   & 0.759   & 0.753  & 0.085   & 0.689   & 0.661  & 0.141   & 0.744   & 0.755  & 0.110  \\
GCPA \cite{chen2020global}          & 0.777   & 0.688  & 0.078   & 0.779   & 0.714   & 0.077   & 0.838   & 0.832  & 0.062   & 0.767   & 0.730  & 0.111   & 0.833   & 0.829  & 0.075  \\
F3Net\cite{wei2020f3net}            & 0.760   & 0.666  & 0.071   & 0.742   & 0.670   & 0.080   & 0.804   & 0.796  & 0.068   & 0.721   & 0.688  & 0.128   & 0.790   & 0.799  & 0.091  \\
MINet \cite{pang2020multi}          & 0.766   & 0.671  & 0.069   & 0.773   & 0.709   & 0.069   & 0.834   & 0.827  & 0.057   & 0.762   & 0.735  & 0.108   & 0.833   & 0.837  & 0.073  \\
GateNet\cite{zhao2020suppress}      & 0.752   & 0.661  & 0.076   & 0.732   & 0.665   & 0.086   & 0.792   & 0.785  & 0.076   & 0.722   & 0.698  & 0.131   & 0.774   & 0.785  & 0.100  \\
CTDNet \cite{CTDNet21}              & 0.778   & 0.704  & 0.074   & 0.766   & 0.708   & 0.079   & 0.828   & 0.828  & 0.061   & 0.750   & 0.722  & 0.118   & 0.816   & 0.825  & 0.080 \\ 
PSG \cite{PSG21}                    & 0.727   & 0.632  & 0.074   & 0.709   & 0.643   & 0.085   & 0.770   & 0.773  & 0.076   & 0.700   & 0.678  & 0.132   & 0.758   & 0.784  & 0.101 \\ 
PGNet \cite{xie2022pyramid}         & 0.797   & 0.723  & 0.062   & 0.806   & 0.765   & 0.060   & 0.857   & 0.862  & 0.050   & 0.788   & 0.767  & 0.096   & 0.856   & 0.867 & 0.063 \\  
EDN \cite{wu2022edn}                & 0.761   & 0.674  & 0.076   & 0.745   & 0.676   & 0.086   & 0.802   & 0.796  & 0.071   & 0.724   & 0.692  & 0.130   & 0.801   & 0.812  & 0.090 \\
ICON \cite{zhuge2022salient}        & 0.707   & 0.622  & 0.084   & 0.670   & 0.607   & 0.097   & 0.715   & 0.701  & 0.100   & 0.665   & 0.637  & 0.148   & 0.709   & 0.713  & 0.122 \\
\midrule[0.5pt]
Ours   
& \textbf{0.815}   & \textbf{0.750}  & \textbf{0.057}   
& \textbf{0.832}   & \textbf{0.801}  & \textbf{0.053}  
& \textbf{0.875}   & \textbf{0.882}  & \textbf{0.045}   
& \textbf{0.812}   & \textbf{0.795}  & \textbf{0.086}
& \textbf{0.877}   & \textbf{0.893}  & \textbf{0.056} \\ 
\midrule[1.0pt]
\multicolumn{16}{c}{\normalsize Results of setting 2 (The model is trained on compressed data (i.e., C-DUTS-TR))} \\
\midrule[0.8pt]
PoolNet    \cite{liu2019simple}     & 0.817  & 0.745  & 0.061  & 0.844  & 0.807  & 0.054  & 0.888  & 0.890  & 0.045 & 0.817  & 0.797  & 0.090 & 0.887  & 0.898 & 0.057 \\
CPD     \cite{wu2019cascaded}       & 0.805  & 0.729  & 0.069  & 0.822  & 0.772  & 0.064  & 0.875  & 0.872  & 0.050 & 0.807  & 0.786  & 0.097 & 0.873  & 0.879 & 0.064 \\
BASNet \cite{qin2019basnet}         & 0.789  & 0.721  & 0.072  & 0.781  & 0.727  & 0.077  & 0.841  & 0.847  & 0.057 & 0.756  & 0.738  & 0.114 & 0.839  & 0.845 & 0.071 \\
EGNet  \cite{zhao2019egnet}         & 0.824  & 0.759  & 0.062  & 0.848  & 0.814  & 0.055  & 0.892  & 0.897  & 0.044 & 0.828  & 0.786  & 0.086 & 0.890  & 0.901 & 0.057 \\
SRCN  \cite{wu2019stacked}          & 0.814  & 0.747  & 0.070  & 0.840  & 0.804  & 0.061  & 0.890  & 0.890  & 0.049 & 0.829  & 0.809  & 0.089 & 0.892  & 0.901 & 0.059 \\
RAS   \cite{chen2020reverse}        & 0.811  & 0.746  & 0.065  & 0.835  & 0.799  & 0.056  & 0.883  & 0.886  & 0.043 & 0.818  & 0.806  & 0.087 & 0.880  & 0.893 & 0.056 \\
GCPA  \cite{chen2020global}         & 0.799  & 0.729  & 0.088  & 0.821  & 0.775  & 0.077  & 0.881  & 0.878  & 0.057 & 0.812  & 0.788  & 0.106 & 0.880  & 0.878 & 0.070 \\
F3Net \cite{wei2020f3net}           & 0.814  & 0.753  & 0.066  & 0.846  & 0.815  & 0.053  & 0.894  & 0.900  & 0.039 & 0.827  & 0.811  & 0.082 & 0.887  & 0.896 & 0.053 \\
MINet \cite{pang2020multi}          & 0.812  & 0.740  & 0.059  & 0.844  & 0.810  & 0.050  & 0.887  & 0.891  & 0.041 & 0.821  & 0.807  & 0.083 & 0.887  & 0.899 & 0.052 \\
GateNet\cite{zhao2020suppress}      & 0.813  & 0.743  & 0.066  & 0.844  & 0.808  & 0.056  & 0.889  & 0.891  & 0.046 & 0.826  & 0.810  & 0.088 & 0.888  & 0.896 & 0.057 \\
CTDNet \cite{CTDNet21}              & 0.816  & 0.757  & 0.059  & 0.848  & 0.822  & 0.047  & 0.892  & 0.902  & 0.039 & 0.825  & 0.818  & 0.080 & 0.886  & 0.903 & 0.051 \\
PSG \cite{PSG21}                    & 0.803  & 0.737  & 0.066  & 0.827  & 0.791  & 0.058  & 0.881  & 0.886  & 0.042 & 0.813  & 0.801  & 0.089 & 0.878  & 0.890 & 0.055 \\
PGNet \cite{xie2022pyramid}         & 0.821  & 0.757  & 0.060  & 0.849  & 0.819  & 0.050  & 0.892  & 0.898  & 0.040 & 0.832  & 0.817  & 0.077 & 0.890  & 0.901 & 0.051 \\
EDN \cite{wu2022edn}                & 0.796  & 0.753  & 0.064  & 0.811  & 0.803  & 0.059  & 0.861  & 0.891  & 0.051 & 0.804  & 0.810  & 0.093 & 0.865  & 0.901 & 0.065 \\
ICON \cite{zhuge2022salient}        & 0.809  & 0.743  & 0.062  & 0.827  & 0.797  & 0.055  & 0.866  & 0.873  & 0.049 & 0.809  & 0.797  & 0.092 & 0.863  & 0.876 & 0.065 \\
\midrule[0.5pt]
Ours   
& \textbf{0.836}   & \textbf{0.784}  & \textbf{0.054}  
& \textbf{0.861}   & \textbf{0.841}  & \textbf{0.044}
& \textbf{0.900}   & \textbf{0.908}  & \textbf{0.036}   
& \textbf{0.840}   & \textbf{0.832}  & \textbf{0.074}
& \textbf{0.897}   & \textbf{0.911}  & \textbf{0.047} \\
\bottomrule[1pt]
\end{tabular}}
\label{table_SOTA_compressed}
\end{table*}

\begin{table*}[!t]
\renewcommand{\arraystretch}{1.}
\centering
\normalsize
\caption{Performances of our approach and state-of-the-art methods (ResNet-based) using S-measure $ (S_m)$ \cite{fan2017structure}, mean absolute error $(M)$, maximum F-measure $ (F_\beta) $ \cite{achanta2009frequency} on clean images (256 $\times$ 256). The best results are shown in \textbf{bold}. Here, $\uparrow$ and $\downarrow$ denote larger and smaller are better, respectively. Our approach can achieve competitive performances in different settings.}
\label{table_SOTA_Clean}
\scalebox{0.75}{
\begin{tabular}{c|p{0.9cm}<{\centering}|p{0.9cm}<{\centering}|p{0.9cm}<{\centering}|p{0.9cm}<{\centering}|p{0.9cm}<{\centering}|p{0.9cm}<{\centering}|p{0.9cm}<{\centering}|p{0.9cm}<{\centering}|p{0.9cm}<{\centering}|p{0.9cm}<{\centering}|p{0.9cm}<{\centering}|p{0.9cm}<{\centering}|p{0.9cm}<{\centering}|p{0.9cm}<{\centering}|p{0.9cm}<{\centering}}
\toprule[1pt]
\multirow{2}{*}{Method} 
& \multicolumn{3}{c|}{DUT-OMRON}  
& \multicolumn{3}{c|}{DUTS-TE} 
& \multicolumn{3}{c|}{HKU-IS}  
& \multicolumn{3}{c|}{PASCAL-S} 
& \multicolumn{3}{c }{ECSSD} \\
& $S_m \uparrow$    & $F_\beta \uparrow$   & $M \downarrow$    
& $S_m \uparrow$    & $F_\beta \uparrow$   & $M \downarrow$ 
& $S_m \uparrow$    & $F_\beta \uparrow$   & $M \downarrow$ 
& $S_m \uparrow$    & $F_\beta \uparrow$   & $M \downarrow$ 
& $S_m \uparrow$    & $F_\beta \uparrow$   & $M \downarrow$ \\
\midrule[0.8pt]
\multicolumn{16}{c}{\normalsize Results of setting 1 (The model is trained on clean data (i.e., DUTS-TR) and then evaluated on various benchmarks)} \\
\midrule[0.8pt]
PoolNet \cite{liu2019simple}        & 0.823  & 0.747   & 0.057  & 0.857  & 0.816  & 0.047  & 0.893  & 0.893  & 0.041  & 0.846  & 0.824  & 0.071  & 0.908  & 0.917  & 0.044          \\
CPD \cite{wu2019cascaded}           & 0.819  & 0.744   & 0.057  & 0.856  & 0.822  & 0.047  & 0.898  & 0.902  & 0.036  & 0.837  & 0.820  & 0.076  & 0.904  & 0.914  & 0.044          \\
BASNet \cite{qin2019basnet}         & 0.834  & 0.780   & 0.056  & 0.858  & 0.833  & 0.049  & 0.902  & 0.915  & 0.034  & 0.830  & 0.824  & 0.079  & 0.903  & 0.921  & 0.042          \\
EGNet\cite{zhao2019egnet}           & 0.817  & 0.734   & 0.056  & 0.854  & 0.811  & 0.047  & 0.892  & 0.888  & 0.041  & 0.837  & 0.803  & 0.074  & 0.906  & 0.909  & 0.046          \\
SRCN \cite{wu2019stacked}           & 0.833  & 0.767   & 0.057  & 0.874  & 0.850  & 0.043  & 0.912  & 0.916  & 0.035  & 0.862  & 0.850  & 0.067  & 0.915  & 0.927  & 0.044          \\
RAS \cite{chen2020reverse}          & 0.834  & 0.776   & 0.053  & 0.872  & 0.850  & 0.041  & 0.909  & 0.916  & 0.032  & 0.849  & 0.842  & 0.070  & 0.909  & 0.923  & 0.042          \\
GCPA \cite{chen2020global}          & 0.836  & 0.771   & 0.057  & 0.882  & 0.858  & 0.040  & 0.915  & 0.921  & 0.033  & 0.858  & 0.844  & 0.065  & 0.917 & 0.927  & 0.039          \\
F3Net \cite{wei2020f3net}           & 0.832  & 0.767   & 0.054  & 0.875  & 0.853  & 0.040  & 0.910  & 0.917  & 0.030  & 0.851  & 0.837  & 0.066  & 0.913  & 0.925  & 0.039          \\
MINet \cite{pang2020multi}          & 0.830  & 0.764   & 0.055  & 0.877  & 0.854  & 0.039  & 0.912  & 0.918  & 0.031  & 0.852  & 0.842  & 0.066  & 0.915  & 0.929  & 0.038          \\
GateNet\cite{zhao2020suppress}      & 0.837  & 0.777   & 0.055  & 0.880  & 0.860  & 0.041  & 0.913  & 0.918  & 0.034  & 0.857  & 0.847  & 0.069  & 0.914  & 0.928  & 0.042          \\
CTDNet\cite{CTDNet21}               & 0.839  & 0.787   & 0.054  & 0.881  & 0.864  & 0.038  & 0.915  & 0.926  & 0.029  & 0.857  & 0.852  & 0.064  & 0.911  & 0.926  & 0.038          \\
PSG\cite{PSG21}                     & 0.829  & 0.769   & 0.053  & 0.873  & 0.853  & 0.039  & 0.908  & 0.919  & 0.031  & 0.851  & 0.846  & 0.066  & 0.910  & 0.927  & 0.038          \\
PGNet\cite{xie2022pyramid}          & 0.838  & 0.778   & 0.053  & 0.886  & 0.871  & 0.036  & 0.916  & 0.927  & 0.029  & 0.858  & 0.848  & 0.063  & 0.914  & 0.928  & 0.038          \\
EDN\cite{wu2022edn}                 & \textbf{0.845}  & \textbf{0.793}   & \textbf{0.051}  & 0.880  & 0.861  & 0.040  & 0.917  & 0.925  & 0.029  & 0.858  & 0.856  & 0.066  & \textbf{0.918} & \textbf{0.932}  & \textbf{0.036}          \\
ICON\cite{zhuge2022salient}         & 0.836  & 0.787   & 0.057  & 0.871  & 0.858  & 0.043  & 0.900  & 0.914  & 0.036  & 0.853  & 0.851 & 0.069  & 0.900  & 0.921  & 0.047          \\
\midrule[0.5pt]
Ours  
& {0.843} & {0.790} & \textbf{0.051} 
& \textbf{0.889} & \textbf{0.875} & \textbf{0.034} 
& \textbf{0.920} & \textbf{0.930} & \textbf{0.028} 
& \textbf{0.867} & \textbf{0.858} & \textbf{0.058} 
& {0.915} & {0.930} & {0.037} \\

\midrule[1.0pt]
\multicolumn{16}{c}{\normalsize Results of setting 2 (The model is trained on compressed data (i.e., C-DUTS-TR) and then evaluated on various benchmarks)} \\
\midrule[0.8pt]
PoolNet\cite{liu2019simple}         & 0.827 & 0.760 & 0.058 & 0.865 & 0.836 & 0.047 & 0.904 & 0.908 & 0.038 & 0.841 & 0.828 & 0.077 & 0.905 & 0.918 & 0.047          \\
CPD \cite{wu2019cascaded}           & 0.817 & 0.746 & 0.067 & 0.847 & 0.806 & 0.056 & 0.894 & 0.895 & 0.042 & 0.829 & 0.808 & 0.086 & 0.894 & 0.903 & 0.055          \\
BASNet \cite{qin2019basnet}         & 0.815 & 0.753 & 0.068 & 0.830 & 0.787 & 0.063 & 0.888 & 0.894 & 0.040 & 0.799 & 0.781 & 0.098 & 0.886 & 0.898 & 0.051          \\
EGNet \cite{zhao2019egnet}          & 0.834 & 0.773 & 0.060 & 0.869 & 0.842 & 0.047 & 0.910 & 0.915 & 0.037 & 0.850 & 0.837 & 0.074 & 0.908 & 0.919 & 0.048          \\
SRCN \cite{wu2019stacked}           & 0.825 & 0.764 & 0.068 & 0.864 & 0.838 & 0.052 & 0.909 & 0.913 & 0.041 & 0.851 & 0.837 & 0.077 & 0.909 & 0.921 & 0.050          \\
RAS \cite{chen2020reverse}          & 0.820 & 0.759 & 0.063 & 0.860 & 0.831 & 0.049 & 0.902 & 0.907 & 0.036 & 0.841 & 0.832 & 0.076 & 0.900 & 0.913 & 0.047          \\
GCPA\cite{chen2020global}          & 0.806  & 0.741 & 0.086 & 0.836 & 0.796 & 0.072 & 0.894 & 0.893 & 0.051 & 0.823 & 0.804 & 0.101 & 0.891 & 0.896 & 0.065          \\
F3Net \cite{wei2020f3net}           & 0.828 & 0.771 & 0.062 & 0.873 & 0.853 & 0.044 & 0.913 & 0.920 & 0.032 & 0.848 & 0.834 & 0.071 & 0.907 & 0.918 & 0.044          \\
MINet\cite{pang2020multi}           & 0.824 & 0.756 & 0.057 & 0.870 & 0.846 & 0.042 & 0.906 & 0.912 & 0.034 & 0.845 & 0.834 & 0.072 & 0.904 & 0.916 & 0.043          \\
GateNet\cite{zhao2020suppress}      & 0.821 & 0.754 & 0.065 & 0.868 & 0.842 & 0.048 & 0.908 & 0.912 & 0.038 & 0.848 & 0.834 & 0.076 & 0.905 & 0.914 & 0.048          \\
CTDNet\cite{CTDNet21}           & 0.829  & 0.771   & 0.053  & 0.877  & 0.858  & 0.038  & 0.913  & 0.922  & \textbf{0.030}  & 0.850  & \textbf{0.845}  & 0.067  & 0.906  & 0.923  & 0.042          \\
PSG\cite{PSG21}      & 0.816  & 0.753   & 0.062  & 0.855  & 0.827  & 0.048  & 0.902  & 0.908  & 0.035  & 0.833  & 0.823  & 0.078  & 0.897  & 0.910  & 0.047          \\
PGNet\cite{xie2022pyramid}      & 0.829  & 0.769   & 0.059  & 0.869  & 0.847  & 0.044  & 0.907  & 0.912  & 0.035  & 0.852  & 0.841  & 0.067  & 0.906  & 0.918  & 0.044          \\
EDN\cite{wu2022edn}      & 0.805  & 0.768   & 0.062  & 0.835  & 0.837  & 0.050  & 0.878  & 0.912  & 0.044  & 0.831  & 0.837  & 0.079  & 0.886  & 0.921  & 0.054          \\
ICON\cite{zhuge2022salient}      & 0.826  & 0.768   & 0.057  & 0.861  & 0.843  & 0.045  & 0.893  & 0.904  & 0.040  & 0.834  & 0.824 & 0.081  & 0.887  & 0.902  & 0.054          \\
\midrule[0.5pt]
Ours          
& \textbf{0.845} & \textbf{0.796} & \textbf{0.052} 
& \textbf{0.882} & \textbf{0.867} & \textbf{0.037} 
& \textbf{0.915} & \textbf{0.924} & \textbf{0.030} 
& \textbf{0.856} & \textbf{0.845} & \textbf{0.065} 
& \textbf{0.911} & \textbf{0.924} & \textbf{0.041} \\
\bottomrule[1pt]
\end{tabular}}
\end{table*}

\section{Experiments}\label{Experiments}
\subsection{Benchmarks and Evaluation Metrics}
\textbf{Compressed Image Benchmarks.} 
To comprehensively evaluate the robustness of SOD methods on compressed images, we perform the acknowledged HEVC-based image compression \cite{sullivan2012overview} on the official SOD testing and training benchmarks. The level of compression is determined by the quantization step and higher quantization parameter (QP) indicates severer degradation in images. 
1) \textit{Testing}: The robust SOD benchmarks contain 5 datasets: C-DUTS-TE, C-DUT-OMRON, C-HKU-IS, C-PASCAL-S and C-ECSSD, which are constructed by performing compression with 5 typical degradation levels (i.e., QP22, QP27, QP32, QP37, and QP42) on DUTS-TE (5019 samples) \cite{wang2017learning}, DUT-OMRON (5168 samples) \cite{yang2013saliency}, HKU-IS (4447 samples) \cite{li2015visual}, PASCAL-S (850 samples) \cite{li2014secrets} and ECSSD (1000 samples) \cite{yan2013hierarchical}, respectively. 
Hence, the total number of each testing benchmark is 5 times the original testing dataset. 
2) \textit{Training}: We perform compression with 5 above-mentioned degradation levels on DUTS-TR (10553 samples) \cite{wang2017learning} to construct the training benchmark, i.e., C-DUTS-TR. It contains the same number as DUTS-TR, and the degradation level of each image is random.

\textbf{Evaluation Protocol.}
We adopt three recognized metrics for evaluation, including S-measure $(S_m)$ \cite{fan2017structure}, F-measure $(F_\beta)$ \cite{achanta2009frequency} and mean absolute error $(M)$. Specifically, $S_m$ evaluates region and object structural similarities. $F_\beta$ can evaluate the overall performance of the saliency map and we provide the results of {maximum} F-measure for comparison. $M$ measures the average absolute difference.

\subsection{Experimental Setup} 
\textbf{Model Settings. }
Our model is conducted on a single Tesla V100 GPU with PyTorch. ResNet50 \cite{he2016deep}, pre-trained on ImageNet, is employed as our backbone, and other parameters are randomly initialized in the prior network and the target network. Note that the prior network and target network have the same structure. Our target network aims to achieve robust compressed image salient object detection performance by employing the proposed hybrid prior learning strategy.

\textbf{Training and Inference Setting. } \label{Experimental Setup}
Adam optimizer is used to train our method for 100 epochs with a batch size of 12. A linear decay learning rate scheme is adopted in our experiment and the maximum learning rate is set as 1e-4. 
The images of training and testing benchmarks are uniformly resized to 256 $\times$ 256.
In this work, we define two comparison settings to comprehensively evaluate the robustness of SOD models. 
\textit{\textbf{Setting 1}}: The model is trained on clean data (i.e., DUTS-TR) and then evaluated on various benchmarks. 
\textit{\textbf{Setting 2}}: The model is trained on compressed data (i.e., C-DUTS-TR) and then evaluated on various benchmarks. 
Therefore, we employ DUTS-TR to train the prior network and C-DUTS-TR to train our target network. In evaluations, we simultaneously analyze the robustness of the prior network and the target network for comprehensive evaluations.

\subsection{Comparison with the State-of-the-arts}
We compare our approach with 15 state-of-the-art CNNs-based SOD models, including PoolNet \cite{liu2019simple}, CPD \cite{wu2019cascaded}, BASNet \cite{qin2019basnet}, EGNet \cite{zhao2019egnet}, SRCN \cite{wu2019stacked}, RAS \cite{chen2020reverse}, GCPA \cite{chen2020global}, F3Net \cite{wei2020f3net}, MINet \cite{pang2020multi}, GateNet \cite{zhao2020suppress}, CTDNet \cite{CTDNet21}, PSG \cite{PSG21}, PGNet \cite{xie2022pyramid}, EDN \cite{wu2022edn} and ICON \cite{zhuge2022salient}. For fair comparisons, the saliency results of these methods are generated by authorized codes with default settings.  

\textbf{Quantitative Analysis on Compressed Images.}  \label{Quantitative Analysis}
In Table \ref{table_SOTA_compressed}, we can observe that our approach can obtain superior performance and outperform other competitors, including setting 1 (The model is trained on clean data and evaluated on compressed data) and setting 2 (The model is trained on compressed data and evaluated on compressed data). 
Specifically, 
1) when training on compressed data (i.e., setting 2) and then evaluating on compressed images, our proposed method exceeds other state-of-the-art models on all challenging datasets. It shows the effectiveness and robustness of our framework when confronted with challenging compressed data. 
2) Importantly, when only training on clean data (i.e., setting 1) and then evaluating on {unforeseen compressed images with unknown degradation levels}, our approach remarkably outperforms the previous methods on different challenging datasets. On the C-DUT-OMRON dataset and C-DUTS-TE dataset, our model obtains large improvements of 3.7\% and 4.6\% in terms of the F-measure, respectively. 
These better results of our approach verify that it is more robust to \textit{unpredictable} compressed data. This indicates that our method has greater confidence in robust compressed image salient object detection in the real world.

\textbf{Quantitative Analysis on Clean Images.}
Moreover, in Table \ref{table_SOTA_Clean}, we report the comparisons between our proposed model and other state-of-the-art approaches on clean images. Particularly, consistent with the size of the compressed image, the original size of the clear image is 256 $\times$ 256. As can be seen from the results, our approach can obtain competitive performance on clean images under different settings. This validates the effectiveness and robustness of our proposed method by achieving superior performances for compressed images with different degradation levels, and maintaining competitive accuracy on clean images.

\begin{figure*}
\centering
\includegraphics[width=\linewidth]{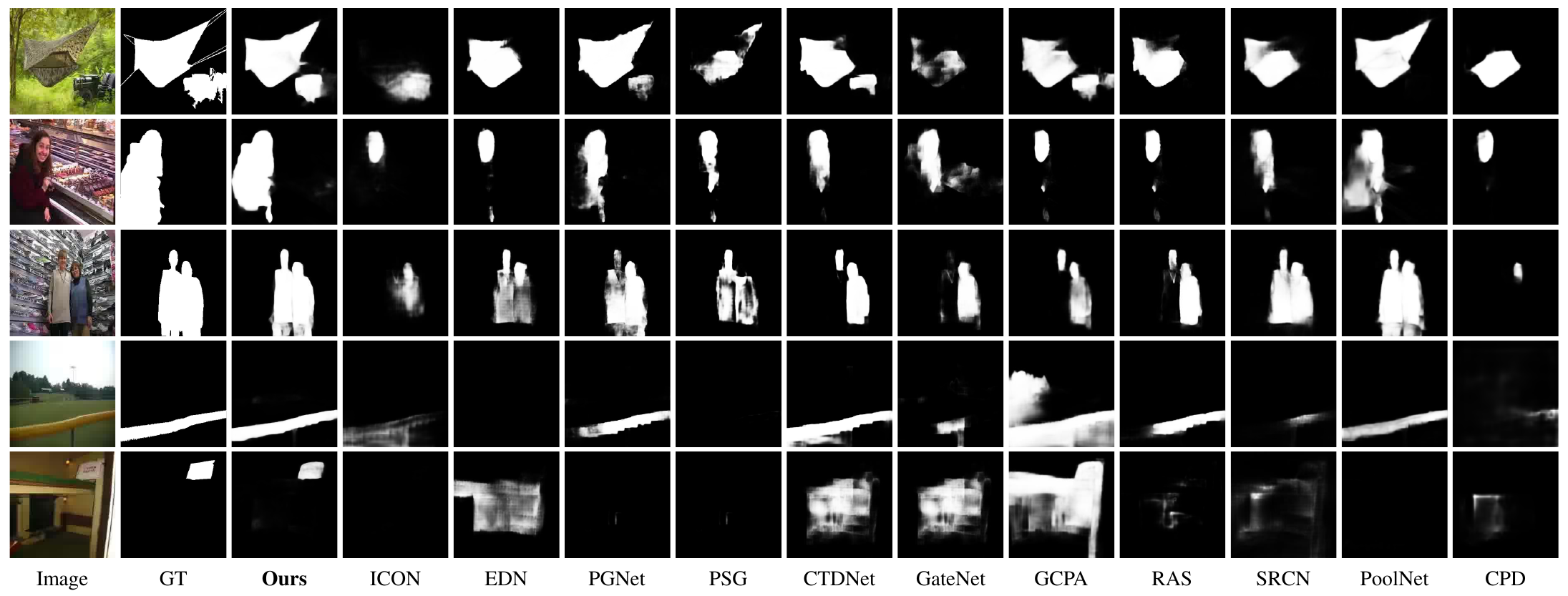}
\caption{
Visual results of setting 1, including our proposed approach and other SOTA methods. From top to bottom, the degradation effect of images gradually increases (i.e., QP22, QP27, QP32, QP37, and QP42), resulting in increased difficulty in detection. In general, it can be observed that our approach can generate more robust and accurate saliency results under different degradation levels.
}
\label{fig:Visual_Comparison_set1}
\end{figure*}

\begin{figure*}
\centering
\includegraphics[width=\linewidth]{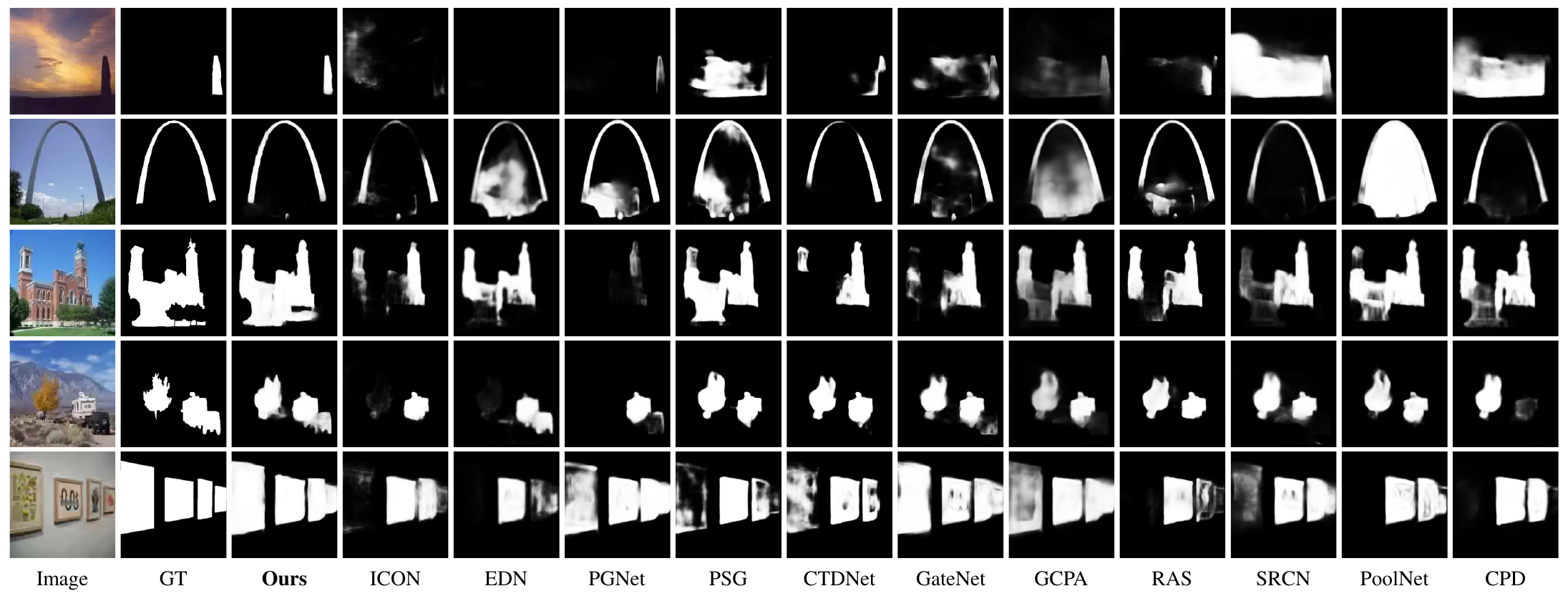}
\caption{
Visual results of setting 2, including our proposed approach and other SOTA methods. From top to bottom, the degradation effect of images gradually increases (i.e., QP22, QP27, QP32, QP37, and QP42), resulting in increased difficulty in detection. In general, it can be observed that our approach can generate more robust and accurate saliency results under different degradation levels.
}
\label{fig:Visual_Comparison}
\end{figure*}

\textbf{Visual Analysis.}
In Fig. \ref{fig:Visual_Comparison_set1} and Fig. \ref{fig:Visual_Comparison}, we provides some visual examples from different degraded compressed images for qualitative comparison against the existing approaches to verify the effectiveness of our method. 
Specifically, the images from the first row to the last row correspond to five degradation levels (i.e., QP22, QP27, QP32, QP37, and QP42). Thus, the distortion level increases from mild to severe from top to bottom. 
In general, benefiting from the effective ability to leverage relationship and location information, it can be seen that our proposed approach generates more complete saliency maps, which are closer to the ground truths. 
Especially, when dealing with slightly degraded images as shown in the first row, e.g., QP22, our method can accurately locate the salient objects. When it comes to seriously degraded scenes as shown in the last row, e.g., QP42, our model can still integrally detect the salient objects. 
These results demonstrate that our approach is more robust in detecting salient objects than existing methods.

\subsection{Ablation Analysis and Discussion}
In this section, we investigate the effectiveness of our proposed components and report the average results of five degradation levels for ablation comparisons. 
Our full model consists of relation prior learning, location prior learning, self-masked learning, and location-aware graph reasoning components. Compared to the full model, we progressively remove our proposed components to investigate their performances.

\begin{table*}[!t]
\renewcommand{\arraystretch}{1.}
\centering
\normalsize
\caption{
Ablation analysis of our approach.  
Here, “w/o” denotes the removing operation. 
“I”:  w/o location prior learning.
“II”: w/o relation prior learning. 
“III”: w/o short-range relation prior learning. 
“IV”: w/o long-range relation prior learning. 
“V”: w/o self-masked learning. 
“VI”: w/o location-aware graph reasoning.  
“Ours”: Baseline + relation prior learning + location prior learning + self-masked learning + location-aware graph reasoning (Full model). 
}
\scalebox{0.90}{
\begin{tabular}{c|p{0.9cm}<{\centering}|p{0.9cm}<{\centering}|p{0.9cm}<{\centering}|p{0.9cm}<{\centering}|p{0.9cm}<{\centering}|p{0.9cm}<{\centering}|p{0.9cm}<{\centering}|p{0.9cm}<{\centering}|p{0.9cm}<{\centering}|p{0.9cm}<{\centering}|p{0.9cm}<{\centering}|p{0.9cm}<{\centering}}
\toprule[1pt]
\multirow{2}{*}{Method} 
& \multicolumn{3}{c|}{C-DUT-OMRON}  
& \multicolumn{3}{c|}{C-DUTS-TE} 
& \multicolumn{3}{c|}{C-HKU-IS}  
& \multicolumn{3}{c }{C-ECSSD} \\
& $S_m \uparrow$    & $F_\beta \uparrow$   & $M \downarrow$    
& $S_m \uparrow$    & $F_\beta \uparrow$   & $M \downarrow$ 
& $S_m \uparrow$    & $F_\beta \uparrow$   & $M \downarrow$ 
& $S_m \uparrow$    & $F_\beta \uparrow$   & $M \downarrow$ \\
\midrule[0.8pt]
I       & 0.830   & 0.775  & 0.060   & 0.854   & 0.831   & 0.049   & 0.894   & 0.901  & 0.040   & 0.893  & 0.905  & 0.050 \\
II      & 0.829   & 0.773  & 0.059   & 0.854   & 0.829   & 0.047   & 0.894   & 0.901  & 0.039   & 0.891   & 0.904  & 0.050 \\
III     & 0.833   & 0.781  & 0.056   & 0.855   & 0.832   & 0.047   & 0.897   & 0.906  & 0.037   & 0.894   & 0.909  & 0.048 \\
IV      & 0.833   & 0.778  & 0.057   & 0.859   & 0.837   & 0.046   & 0.899   & 0.906  & 0.038   & 0.895   & 0.908  & 0.049 \\
V       & 0.830   & 0.779  & 0.056   & 0.859   & 0.839   & 0.046   & 0.899   & 0.907  & 0.036   & 0.896   & 0.911  & 0.047 \\
VI      & 0.831   & 0.774  & 0.054   & 0.858   & 0.835   & 0.045   & 0.897   & 0.905  & 0.037   & 0.891   & 0.905  & 0.049 \\
\midrule[0.5pt]
Ours
& \textbf{0.836}   & \textbf{0.784}  & \textbf{0.054}  
& \textbf{0.861}   & \textbf{0.841}  & \textbf{0.044}
& \textbf{0.900}   & \textbf{0.908}  & \textbf{0.036}   
& \textbf{0.897}   & \textbf{0.911}  & \textbf{0.047} \\
\bottomrule[1pt]
\end{tabular}}
\label{table_ablation_FullModel}
\end{table*}


\subsubsection{Effectiveness of Different Components}
In Table \ref{table_ablation_FullModel}, we gradually remove different components (ID: I-VI) in comparison with our full model (ID: Ours). 
As shown in Table \ref{table_ablation_FullModel}, we first test the location prior learning strategy (ID: I and Ours), which shows a consistent performance gain on different datasets. 
This suggests that location prior learning can facilitate our network to make good use of the location information and thus improve the robustness of detection. 
Besides, when applying the relation prior learning (ID: II, III, IV, and Ours), our method again demonstrates the improvement. 
This reveals that the relation prior learning can promote our network to well leverage the relationship information, which is beneficial for network robustness. 
With the cooperation of self-masked learning (ID: V and Ours), the accuracy can be improved. 
Finally, we introduce the proposed location-aware graph reasoning (ID: VI and Ours) to generate the saliency prediction. We can find that our LGR can effectively improve the model performance by better capturing the object parts of salient objects. 
In general, the quantitative results of applying our HPL and LGR validate the advantages of compressed image salient object detection.

\begin{table*}[!t]
\renewcommand{\arraystretch}{1.}
\centering
\normalsize
\caption{
Ablation analysis of our hybrid prior learning with different settings. Here, “w/” and “w/o” denote the replacing operation and the removing operation, respectively.
“I”: w/o foreground-guided mask in the location prior learning. 
“II”: w/o relation matrix in the long-range relation prior learning. 
“III”: w/ MAE loss instead of using structure learning loss.  
“IV”: w/ MSE loss instead of using structure learning loss.  
“Ours”: Full model. 
}
\scalebox{0.90}{
\begin{tabular}{c|p{0.9cm}<{\centering}|p{0.9cm}<{\centering}|p{0.9cm}<{\centering}|p{0.9cm}<{\centering}|p{0.9cm}<{\centering}|p{0.9cm}<{\centering}|p{0.9cm}<{\centering}|p{0.9cm}<{\centering}|p{0.9cm}<{\centering}|p{0.9cm}<{\centering}|p{0.9cm}<{\centering}|p{0.9cm}<{\centering}}
\toprule[1pt]
\multirow{2}{*}{Method} 
& \multicolumn{3}{c|}{C-DUT-OMRON}  
& \multicolumn{3}{c|}{C-DUTS-TE} 
& \multicolumn{3}{c|}{C-HKU-IS}  
& \multicolumn{3}{c }{C-ECSSD} \\
& $S_m \uparrow$    & $F_\beta \uparrow$   & $M \downarrow$    
& $S_m \uparrow$    & $F_\beta \uparrow$   & $M \downarrow$ 
& $S_m \uparrow$    & $F_\beta \uparrow$   & $M \downarrow$ 
& $S_m \uparrow$    & $F_\beta \uparrow$   & $M \downarrow$ \\
\midrule[0.8pt]
I   & 0.835   & 0.781  & 0.057   & 0.859   & 0.838   & 0.046   & 0.897   & 0.905  & 0.038   & 0.894   & 0.907  & 0.049 \\
II         & 0.833   & 0.780  & 0.055   & 0.861   & 0.839   & 0.044   & 0.899   & 0.907  & 0.037   & 0.894   & 0.907  & 0.048 \\
III        & 0.835   & 0.782  & 0.056   & 0.858   & 0.837   & 0.046   & 0.897   & 0.906  & 0.038  & 0.892   & 0.906  & 0.050 \\
IV         & 0.833   & 0.777  & 0.055   & 0.860   & 0.840   & 0.045   & 0.898   & 0.905  & 0.037   & 0.893   & 0.906  & 0.049 \\
\midrule[0.5pt]
Ours
& \textbf{0.836}   & \textbf{0.784}  & \textbf{0.054}  
& \textbf{0.861}   & \textbf{0.841}  & \textbf{0.044}
& \textbf{0.900}   & \textbf{0.908}  & \textbf{0.036}   
& \textbf{0.897}   & \textbf{0.911}  & \textbf{0.047} \\
\bottomrule[1pt]
\end{tabular}}
\label{table_ablation_HPL}
\end{table*}

\begin{table*}[!t]
\renewcommand{\arraystretch}{1.}
\centering
\normalsize
\caption{
Ablation analysis of our location-aware graph reasoning with different settings. Here, “w/” and “w/o” denote the replacing operation and the removing operation, respectively.
“I”: w/o the connection module. 
“II”: w/o graph reasoning (w channel concat operation). 
“III”: w cross-attention. 
“Ours”: Full model. 
}
\scalebox{0.90}{
\begin{tabular}{c|p{0.9cm}<{\centering}|p{0.9cm}<{\centering}|p{0.9cm}<{\centering}|p{0.9cm}<{\centering}|p{0.9cm}<{\centering}|p{0.9cm}<{\centering}|p{0.9cm}<{\centering}|p{0.9cm}<{\centering}|p{0.9cm}<{\centering}|p{0.9cm}<{\centering}|p{0.9cm}<{\centering}|p{0.9cm}<{\centering}}
\toprule[1pt]
\multirow{2}{*}{Method} 
& \multicolumn{3}{c|}{C-DUT-OMRON}  
& \multicolumn{3}{c|}{C-DUTS-TE} 
& \multicolumn{3}{c|}{C-HKU-IS}  
& \multicolumn{3}{c }{C-ECSSD} \\
& $S_m \uparrow$    & $F_\beta \uparrow$   & $M \downarrow$    
& $S_m \uparrow$    & $F_\beta \uparrow$   & $M \downarrow$ 
& $S_m \uparrow$    & $F_\beta \uparrow$   & $M \downarrow$ 
& $S_m \uparrow$    & $F_\beta \uparrow$   & $M \downarrow$ \\
\midrule[0.8pt]
I   & 0.814   & 0.756  & 0.063   & 0.835   & 0.805   & 0.054   & 0.884  & 0.890  & 0.044   & 0.878   & 0.890  & 0.057 \\
II         & 0.827   & 0.769  & 0.058   & 0.858   & 0.835   & 0.047   & 0.896   & 0.904  & 0.038   & 0.890   & 0.905  & 0.049 \\
III         & 0.830   & 0.774  & 0.054   & 0.859   & 0.837   & 0.044   & 0.898   & 0.907  & 0.037   & 0.894   & 0.910  & 0.047 \\
\midrule[0.5pt]
Ours
& \textbf{0.836}   & \textbf{0.784}  & \textbf{0.054}  
& \textbf{0.861}   & \textbf{0.841}  & \textbf{0.044}
& \textbf{0.900}   & \textbf{0.908}  & \textbf{0.036}   
& \textbf{0.897}   & \textbf{0.911}  & \textbf{0.047} \\
\bottomrule[1pt]
\end{tabular}}
\label{table_ablation_LGR}
\end{table*}

\begin{table*}[!t]
\renewcommand{\arraystretch}{1.}
\centering
\normalsize
\caption{
Comparison analysis of the hyper-parameters for our hybrid prior learning strategy (refer to Equation \eqref{eq_weight_HPL}). 
}
\scalebox{0.90}{
\begin{tabular}{cc|p{0.9cm}<{\centering}|p{0.9cm}<{\centering}|p{0.9cm}<{\centering}|p{0.9cm}<{\centering}|p{0.9cm}<{\centering}|p{0.9cm}<{\centering}|p{0.9cm}<{\centering}|p{0.9cm}<{\centering}|p{0.9cm}<{\centering}|p{0.9cm}<{\centering}|p{0.9cm}<{\centering}|p{0.9cm}<{\centering}}
\toprule[1pt]
\multirow{2}{*}{$\alpha$} & \multirow{2}{*}{$\beta$} 
& \multicolumn{3}{c|}{C-DUT-OMRON}  
& \multicolumn{3}{c|}{C-DUTS-TE} 
& \multicolumn{3}{c|}{C-HKU-IS}  
& \multicolumn{3}{c }{C-ECSSD} \\
& & $S_m \uparrow$    & $F_\beta \uparrow$   & $M \downarrow$    
& $S_m \uparrow$    & $F_\beta \uparrow$   & $M \downarrow$ 
& $S_m \uparrow$    & $F_\beta \uparrow$   & $M \downarrow$ 
& $S_m \uparrow$    & $F_\beta \uparrow$   & $M \downarrow$ \\
\midrule[0.8pt]
0.1 & 0.1  & 0.832 & 0.778 & 0.058 & 0.859 & 0.838 & 0.046 & 0.899 & 0.908 & 0.036 & 0.897   & 0.911 & 0.046 \\
0.5 & 0.5  
& \textbf{0.836}   & \textbf{0.784}  & \textbf{0.054}  
& \textbf{0.861}   & \textbf{0.841}  & \textbf{0.044}
& \textbf{0.900}   & \textbf{0.908}  & \textbf{0.036}   
& \textbf{0.897}   & \textbf{0.911}  & \textbf{0.047} \\
1.0 & 1.0  & 0.834 & 0.780 & 0.056 & 0.861 & 0.840 & 0.044 & 0.900 & 0.909 & 0.035 & 0.896   & 0.910 & 0.046 \\
1.0 & 2.0  & 0.833 & 0.780 & 0.055 & 0.860 & 0.841 & 0.045 & 0.900 & 0.909 & 0.036 & 0.895   & 0.910 & 0.047 \\
2.0 & 1.0  & 0.831 & 0.775 & 0.057 & 0.861 & 0.841 & 0.044 & 0.898 & 0.907 & 0.036 & 0.896   & 0.909 & 0.047 \\
2.0 & 2.0  & 0.830 & 0.777 & 0.058 & 0.861 & 0.840 & 0.045 & 0.900 & 0.909 & 0.036 & 0.895   & 0.911 & 0.047 \\
\bottomrule[1pt]
\end{tabular}}
\label{table_ablation_hyper_parameter}
\end{table*}

\begin{table*}[!t]
\renewcommand{\arraystretch}{1.}
\centering
\normalsize
\caption{
Comparison analysis of the probability of implementing self-masked learning. 
}
\scalebox{0.88}{
\begin{tabular}{c|p{0.9cm}<{\centering}|p{0.9cm}<{\centering}|p{0.9cm}<{\centering}|p{0.9cm}<{\centering}|p{0.9cm}<{\centering}|p{0.9cm}<{\centering}|p{0.9cm}<{\centering}|p{0.9cm}<{\centering}|p{0.9cm}<{\centering}|p{0.9cm}<{\centering}|p{0.9cm}<{\centering}|p{0.9cm}<{\centering}}
\toprule[1pt]
\multirow{2}{*}{Probability} 
& \multicolumn{3}{c|}{C-DUT-OMRON}  
& \multicolumn{3}{c|}{C-DUTS-TE} 
& \multicolumn{3}{c|}{C-HKU-IS}  
& \multicolumn{3}{c }{C-ECSSD} \\
& $S_m \uparrow$    & $F_\beta \uparrow$   & $M \downarrow$    
& $S_m \uparrow$    & $F_\beta \uparrow$   & $M \downarrow$ 
& $S_m \uparrow$    & $F_\beta \uparrow$   & $M \downarrow$ 
& $S_m \uparrow$    & $F_\beta \uparrow$   & $M \downarrow$ \\
\midrule[0.8pt]
5\%   & 0.833 & 0.781 & 0.056 & 0.861 & 0.840 & 0.045 & 0.900 & 0.908 & 0.036 & 0.896 & 0.910 & 0.047 \\
10\%          
& \textbf{0.836}   & \textbf{0.784}  & \textbf{0.054}  
& \textbf{0.861}   & \textbf{0.841}  & \textbf{0.044}
& \textbf{0.900}   & \textbf{0.908}  & \textbf{0.036}   
& \textbf{0.897}   & \textbf{0.911}  & \textbf{0.047} \\
15\%  &  0.834 & 0.779 & 0.057 & 0.859 & 0.836 & 0.046 & 0.900 & 0.907 & 0.037 & 0.897 & 0.911 & 0.047  \\
20\%  &  0.832 & 0.782 & 0.063 & 0.858 & 0.836 & 0.049 & 0.900 & 0.907 & 0.038 & 0.898 & 0.911 & 0.048  \\
25\%  &  0.832 & 0.777 & 0.055 & 0.860 & 0.837 & 0.045 & 0.899 & 0.907 & 0.037 & 0.897 & 0.911 & 0.046  \\
30\%  &  0.833 & 0.776 & 0.059 & 0.856 & 0.831 & 0.048 & 0.898 & 0.904 & 0.038 & 0.898 & 0.910 & 0.047  \\
35\%  &  0.833 & 0.781 & 0.060 & 0.859 & 0.837 & 0.048 & 0.900 & 0.907 & 0.037 & 0.897 & 0.910 & 0.047  \\
40\%  &  0.833 & 0.781 & 0.058 & 0.858 & 0.837 & 0.047 & 0.899 & 0.907 & 0.037 & 0.897 & 0.911 & 0.047  \\
45\%  &  0.829 & 0.779 & 0.062 & 0.854 & 0.832 & 0.049 & 0.897 & 0.905 & 0.039 & 0.896 & 0.911 & 0.047  \\
50\%  &  0.834 & 0.784 & 0.058 & 0.857 & 0.834 & 0.048 & 0.898 & 0.906 & 0.038 & 0.898 & 0.912 & 0.047  \\
\bottomrule[1pt]
\end{tabular}}
\label{table_ablation_mask_ratio}
\end{table*}

\subsubsection{Analysis of Hybrid Prior Learning (HPL)}
To further validate the effectiveness of our HPL, we investigate some variants in Table \ref{table_ablation_HPL}, including removing the foreground-guided mask strategy in the location prior learning (ID: I), removing the relation matrix in the long-range relation prior learning (ID: II), replacing the structure learning loss with MAE loss (ID: III) and replacing the structure learning loss with MSE loss (ID: IV). 

We can find that 
1) removing the foreground-guided mask strategy will result in worse performance. The potential reason for this could be that learning location prior with the whole feature map introduces some background noise, which could mislead the network to learn wrong information. 
2) Removing the relation matrix incurs a performance penalty, which suggests that modeling the relational matrix is beneficial in capturing long-range dependencies and thus improving robust feature learning. 
3) The structure learning loss can better leverage spatial dependencies and promote our network to achieve better saliency results, which is more suitable for the CI SOD task.

\subsubsection{Analysis of Location-aware Graph Reasoning (LGR)}
One of our core claims is that the better SOD on compressed images should construct the relationship between the object parts and the object location according to the above discussion. 
From Table \ref{table_ablation_FullModel}, we can see that consistent performance improvement can be achieved by applying LGR. 
Moreover, we investigate some variants of the location-aware graph reasoning (LGR) in Table \ref{table_ablation_LGR} to validate the effectiveness of our LGR, including removing connection module (CM) in LGR (ID: I), removing the graph reasoning process and using a channel concatenation operation (ID: II), replacing the graph reasoning process with the cross-attention operation (ID: III) and full model (ID: V). 
The results in Table \ref{table_ablation_LGR} verify the effectiveness of improving robustness by applying the CM and the graph reasoning schemes.

\subsubsection{Impact of the hyper-parameters for the HPL}
We compare the impact of the hyper-parameters by tuning $\alpha$ and $\beta$, respectively (refer to Equation \eqref{eq_weight_HPL}). We study their influence and show the comparison results in Table \ref{table_ablation_hyper_parameter}. 
It can be seen that the effects of setting $\alpha$ and $\beta$ too small (e.g., $\alpha$=0.1, $\beta$=0.1) or too large (e.g., $\alpha$=2.0, $\beta$=2.0) are not optimal, but the overall performance remains relatively stable. 
We find that the performance is better when $\alpha$ and $\beta$ are set to 0.5, respectively. Therefore, we finally adopt this setting, i.e., $\alpha$=0.5 and $\beta$=0.5, in our experiments to obtain better saliency results.

\subsubsection{Impact of the probability of implementing SML}
We study the influence of the probability of implementing our self-masked learning (SML) in Table \ref{table_ablation_mask_ratio}. 
We find that the performance will be unstable with too high implementing probability. 
We infer that the more samples by conducting self-masking, the more difficult the model learns, which in turn is not conducive to the model mining robust features. 
When set at a more appropriate learning difficulty, SML can facilitate the robust saliency feature learning of our model. 
When we set the probability to 10\%, the model achieves a more stable and better performance.

\section{Conclusion}\label{Conclusion}
In this work, we delve into the realm of robust compressed image salient object detection (CI SOD) issue, which is a challenging yet underexplored area. 
Our initial step involves a comprehensive evaluation of robust CI SOD, coupled with an in-depth analysis of contemporary SOD models. 
Building upon the observations of the limitations inherent in existing CNN-based SOD methods and the unique characteristics of compressed images, we propose a simple yet promising baseline framework to improve the robustness of CI SOD. 
Our framework takes into account the broken structure and missing information characteristics of compressed images, and designs a hybrid prior learning strategy to address these inherent challenges for robust feature learning within compressed images.
Extensive experiments on various benchmarks validate the effectiveness and robustness of our proposed approach across various degradation levels of compressed images, while maintaining competitive accuracy on clean images. 

In the future, we hope that the compressed image salient object detection task would help the community towards a more comprehensive understanding of the robustness of CNN-based SOD algorithms. We also hope that it will stimulate the interest and inspiration of more researchers.

\bibliographystyle{unsrt}  
\bibliography{CISOD}

\end{document}